\documentclass[acmtog]{acmart}

\usepackage{booktabs} 

\setcopyright{rightsretained}
\acmJournal{TOG}
\acmYear{2018}\acmVolume{37}\acmNumber{4}\acmArticle{65}\acmMonth{8}
\acmDOI{10.1145/3197517.3201323}

\setcitestyle{square}

\usepackage{graphicx}
\usepackage[ruled]{algorithm2e} 

\newcommand{\PDI}{Multiplane Image\xspace}
\newcommand{\Pdi}{Multiplane image\xspace}
\newcommand{\PDIshort}{MPI\xspace}
\newcommand{\aPDIshort}{an MPI\xspace}
\newcommand{\APDIshort}{An MPI\xspace}
\newcommand{\PDIs}{Multiplane Images\xspace}
\newcommand{\PDIshorts}{MPIs\xspace}
\newcommand{\pdi}{multiplane image\xspace}
\newcommand{\pdis}{multiplane images\xspace}
\newcommand{\OrbSlam}{ORB-SLAM2\xspace}

\newcommand{\new}[1]{{#1}}

\SetAlFnt{\small}
\SetAlCapFnt{\small}
\SetAlCapNameFnt{\small}
\SetAlCapHSkip{0pt}
\IncMargin{-\parindent}

\begin{document}
\title{Stereo Magnification: Learning view synthesis using multiplane images
}

\author{Tinghui Zhou}
\affiliation{
  \institution{University of California, Berkeley}
}
\author{Richard Tucker}
\affiliation{
  \institution{Google}
}
\author{John Flynn}
\affiliation{
  \institution{Google}
}
\author{Graham Fyffe}
\affiliation{
  \institution{Google}
}
\author{Noah Snavely}
\affiliation{
  \institution{Google}
}

\begin{abstract}
The view synthesis problem---generating novel views of a scene from known
imagery---has garnered recent attention due in part to compelling applications
in virtual and augmented reality. In this paper, we explore an intriguing
scenario for view synthesis: extrapolating views from imagery captured by
narrow-baseline stereo cameras, including VR cameras and now-widespread
dual-lens camera phones. We call this problem \emph{stereo magnification}, and
propose a learning framework that leverages a new layered representation that we
call \emph{multiplane images} (MPIs). Our method also uses a massive new data
source for learning view extrapolation: online videos on YouTube. Using data
mined from such videos, we train a deep network that predicts an MPI from an
input stereo image pair. This inferred MPI can then be used to synthesize a
range of novel views of the scene, including views that extrapolate
significantly beyond the input baseline. We show that our method compares
favorably with several recent view synthesis methods, and demonstrate
applications in magnifying narrow-baseline stereo images.

\end{abstract}



\begin{teaserfigure}
\centering
\includegraphics[width=\linewidth]{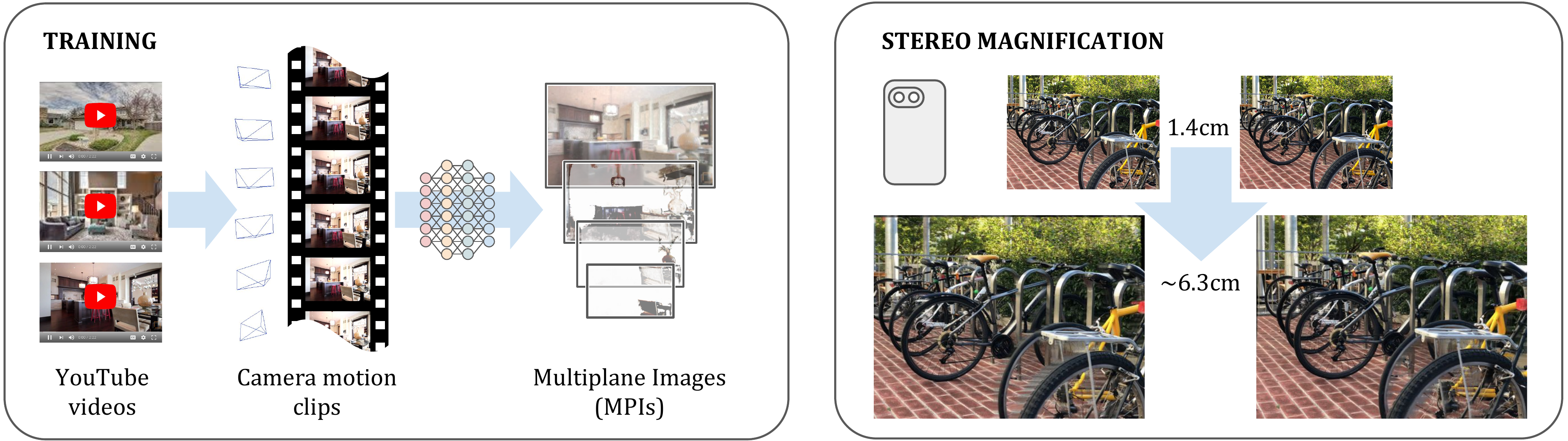}
\caption{We extract camera motion clips from YouTube videos and use them to train a neural network to generate the \PDI (\PDIshort) scene representation from narrow-baseline stereo image pairs. The inferred \PDIshort representation can then be used to synthesize novel views of the scene, including ones that extrapolate significantly beyond the input baseline. (Video stills in this and other figures are used under Creative-Commons license from YouTube user \textit{SonaVisual}.)}
\label{fig:teaser}
\end{teaserfigure}

\maketitle

\section{Introduction}

Photography has undergone an upheaval over the past decade. Cellphone cameras have steadily displaced point-and-shoot cameras, and have become competitive with digital SLRs in certain scenarios. This change has been driven by the increasing image quality of cellphone cameras, through better hardware and also
through computational photography functionality such as high dynamic range imaging~\cite{hasinoff2016burst} and synthetic defocus~\cite{appleportraitmode,googleportrait}. Many of these recent innovations have sought to replicate capabilities of traditional cameras. However, cell phones are also rapidly acquiring new kinds of sensors, such as multiple lenses and depth sensors, enabling applications beyond traditional photography.

In particular, dual-lens cameras are becoming increasingly common. While stereo cameras have been around for nearly as long as photography itself, 
recently a number of dual-camera phones, such as the iPhone 7, have appeared on the market. These cameras tend to have a very small baseline (distance between views) on the order of a centimeter. We have also seen the recent appearance of a number of ``virtual-reality ready'' cameras that capture stereo images and video from a pair of cameras spaced approximately eye-distance apart~\cite{vr180}.

Motivated by the proliferation of stereo cameras, our paper explores the problem
of synthesizing new views from such narrow-baseline image pairs. While much
prior work has explored the problem of \emph{interpolating} between a set of
given views~\cite{chen1992view}, we focus on the problem of \emph{extrapolating}
views significantly beyond the two input images. Such view extrapolation has
many applications for photography. For instance, we might wish to take a
narrow-baseline ($\sim$1cm) stereo pair on a cell phone and extrapolate to an
IPD-separated ($\sim$6.3cm) stereo pair so as to create a photo with a
compelling 3D stereo effect. Or, we might wish to take an IPD-separated stereo
pair captured with a VR180 camera and extrapolate to an entire set of views
along a line say half a meter in length, so as to enable full parallax with a
small range of head motion. We call such view extrapolation from pairs of input
views \emph{stereo magnification}. The examples above involve magnifying the
baseline by a significant amount---up to about 8x the original baseline.

The stereo magnification problem is challenging. We have just two views as
input, unlike in common view interpolation scenarios that consider multiple
views. We wish to be able to handle challenging scenes with reflection and
transparency. Finally, we need the capacity to render pixels that are occluded
and thus not visible in either input view. To address these challenges, our
approach is to \emph{learn} to perform view extrapolation from large amounts of
visual data, following recent work on deep learning for view
interpolation~\cite{flynn2016deepstereo,kalantari2016learning}. However, our
approach differs in key ways from prior work. First, we seek a scene
representation that can be predicted once from a pair of input views, then
reused to predict many output views, unlike in prior work where each output view
must be predicted separately. Second, we need a representation that can
effectively capture surfaces that are hidden in one or both input views. We
propose a layered representation called a \PDI (\PDIshort) that has both of
these properties. Finally, we need training data that matches our task. Simply
collecting stereo pairs is not sufficient, because for training we also require
additional views that are some distance from an input stereo pair as our ground
truth. We propose a simple, surprising source for such data---online video,
e.g., from YouTube, and show that large amounts of suitable data can be mined at
scale for our task.

In experiments we compare our approach to recent view synthesis methods, and perform a number of ablation studies. We show that our method achieves better numerical performance on a held-out test set, and also produces more spatially stable output imagery since our inferred scene representation is shared for synthesizing all target views.
\new{We also show that our learned model generalizes to other datasets without re-training, and is effective at magnifying the narrow baseline of stereo imagery captured by cell phones and stereo cameras.}

In short, our contributions include:
\begin{itemize}
    \item A learning framework for stereo magnification (view extrapolation from narrow-baseline stereo imagery).
    \item \PDIs, a new scene representation for performing view synthesis. 
    \item A new use of online video for learning view synthesis, and in particular view extrapolation.
\end{itemize}

\section{Related work}

\paragraph{Classical approaches to view synthesis}
View synthesis---i.e., taking one or more views of a scene as input, and generating novel views---is a classic problem in computer graphics that forms the core of many image-based rendering systems. Many approaches focus on the interpolation setting, and operate by either interpolating rays from dense imagery (``light field rendering'')~\cite{levoy1996lightfield,gortler1996lumigraph}, or reconstructing scene geometry from sparse views~\cite{debevec1996modeling,zitnick2004interpolation,hedman2017casual}. While these methods yield high-quality novel views, they do so
by compositing the corresponding input pixels/rays, and typically only work well with multiple ($> 2$) input views. \new{View synthesis from stereo imagery has also been considered, including converting 3D stereoscopic video to multi-view video suitable for glasses-free automultiscopic displays~\cite{riechert2012stereo,didyk2013joint,chapiro2014optimizing,kellnhofer20173dtv} and 4D light field synthesis from a micro-baseline stereo pair~\cite{zhang2015light}, as well as generalizations that reconstruct geometry from multiple small-baseline views~\cite{yu2014accidental,ha2016high}.}  While we also focus on stereo imagery, the techniques we present can also be adapted to single-view and multi-view settings. We also target much larger extrapolations than prior work.

\paragraph{Learning-based view synthesis}
More recently, researchers have applied powerful deep learning techniques to view synthesis. View synthesis can be naturally formulated as a learning problem by capturing images of a large number of scenes, withholding some views of each scene as ground truth, training a model that predicts such missing views from one or more given views, and comparing these predicted views to the ground truth as the loss or objective that the learning seeks to optimize. Recent work has explored a number of deep network architectures, scene representations, and application scenarios for learning view synthesis. 

Flynn et al.~\shortcite{flynn2016deepstereo} proposed a view interpolation
method called DeepStereo that predicts a volumetric representation from a set of
input images, and trains a model using images of street scenes. Kalantari et
al.~\shortcite{kalantari2016learning} use light field photos captured by a Lytro
camera~\cite{lytro} as training data for predicting
a color image for a target interpolated viewpoint. Both of these methods predict
a representation in the coordinate system of the \emph{target} view. Therefore,
these methods must run the trained network for each desired target view, making
real-time rendering a challenge. Our method predicts the scene
representation once, and reuses it to render a range of output views in real
time. Further, these prior methods focus on interpolation, rather than
extrapolation as we do.

Other recent work has explored the problem of synthesizing a stereo
pair~\cite{xie2016deep3d}, large camera motion~\cite{zhou2016view}, or even a
light field~\cite{pratul2017lightfield} from a \emph{single} image, an extreme
form of extrapolation. \new{Our work focuses on the increasingly common scenario
  of narrow-baseline stereo pairs. This two-view scenario potentially allows for
  generalization to more diverse scenes and larger extrapolation than the single-view
  scenario.
  The recent single-view method of Srinivasan et al., for instance, only considers
  relatively homogeneous datasets such as macro shots of flowers, and extrapolates
  up to the small baseline of a Lytro camera, whereas our method is able to operate on
  diverse sets of indoor and outdoor scenes, and extrapolate views sufficient to
  allow slight head motions in a VR headset.}

Finally, a variety of work in computer vision has used view synthesis as an
indirect form of supervision for other tasks, such as predicting depth, shape,
or optical flow from one or more
images~\cite{garg2016unsupervised,godard2016unsupervised,zhou2017unsupervised,tulsiani2017multiview,vijayanarasimhan2017sfmnet,liu2017video}. However,
view synthesis is not the explicit goal of such work.



\paragraph{Scene representations for view synthesis}

A wide variety of scene representations have been proposed for modeling scenes in view synthesis tasks. We are most interested in representations that can be predicted once and then reused to render multiple views at runtime. To achieve such a capability, representations are often volumetric or otherwise involve some form of \emph{layering}. For instance, layered depth images (LDIs) are a generalization of depth maps that represent a scene using several layers of depth maps and associated color values~\cite{shade1998layered}. Such layers allow a user to ``see around'' the foreground geometry to the occluded objects that lie behind.
Zitnick et al., represent scenes using per-input-image depth maps, but also solve for alpha matted layers around depth discontinuities to achieve high-quality interpolation~\shortcite{zitnick2004interpolation}. Perhaps closest to our representation is that of Penner and Zhang~\shortcite{penner2017soft3d}. They achieve softness by explicitly modeling confidence, whereas we model transparency which leads to a different method of compositing and rendering. Additionally, whereas we build one representation of a scene, they produce a representation for each input view and then interpolate between them. Our representation is also related to the classic layered representation for encoding moving image sequences by Wang and Adelson~\shortcite{wang1994representing}, and to the layered attenuators of Wetzstein, et al.~\shortcite{wetzstein2011layered}, who use actual physical printed transparencies to construct lightfield displays. \new{Finally, Holroyd et al~\shortcite{holroyd2011multilayer} explore a similar representation to ours but in physical form.}

The \pdi (\PDIshort) representation we use combines several attractive properties of prior methods, including handling of multiple layers and ``softness'' of layering for representing mixed pixels around boundaries or reflective/transparent objects. Crucially, we also found it to be suitable for learning via deep networks.






%

\section{Approach}
Given two images $I_1$ and $I_2$ with known camera parameters, our goal is to learn a deep neural net to infer a global scene representation suitable for synthesizing novel views of the same scene, and in particular extrapolating beyond the input views. In this section, we first describe our scene representation and its characteristics, and then present our pipeline and objective for learning to predict such representation. Note that while we focus on stereo input in this paper, our approach could be adapted to more general view synthesis setups with either single or multiple input views.

\begin{figure}[t]
\begin{center}
\centering
\includegraphics[width=\linewidth]{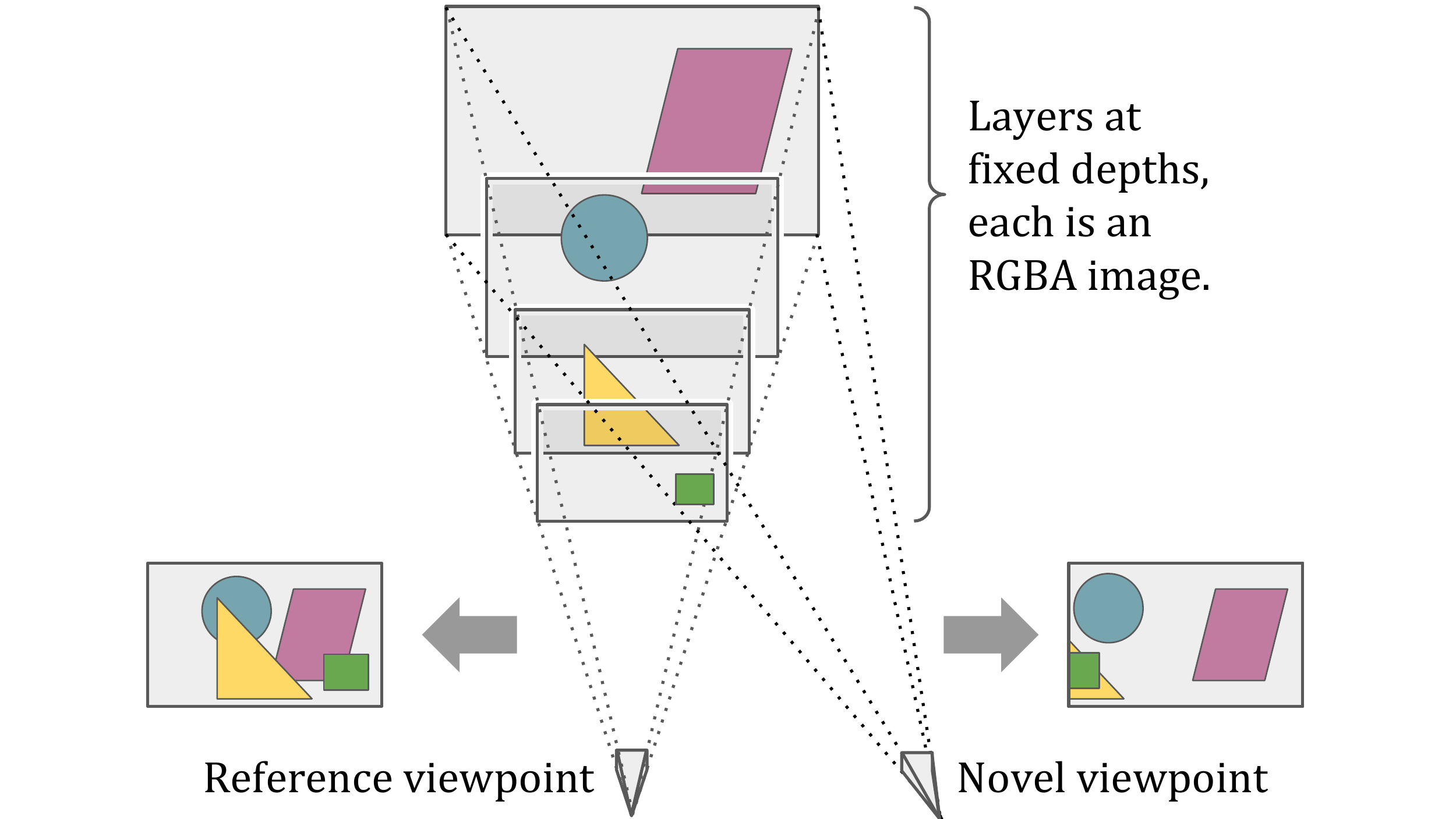}
\caption{\new{An illustration of the \pdi (\PDIshort) representation. An \PDIshort consists of a set of fronto-parallel planes at fixed depths from a reference camera coordinate frame, where each plane encodes an RGB image and an alpha map that capture the scene appearance at the corresponding depth. The \PDIshort representation can be used for efficient and realistic rendering of novel views of the scene.}}
\label{fig:representation}
\end{center}
\end{figure}

\subsection{\Pdi representation}
The global scene representation we adopt is a set of fronto-parallel planes at a fixed range of depths with respect to a reference coordinate frame, where each plane $d$ encodes an RGB color image $C_d$ and an alpha/transparency map $\alpha_d$. Our representation, which we call a \emph{\PDI}~(\PDIshort), can thus be described as a collection of such RGBA layers $\{(C_1, \alpha_1), \ldots, (C_D, \alpha_D)\}$, where $D$ is the number of depth planes. \APDIshort is related to the \emph{Layered Depth Image} (LDI) representation of Shade, et al.~\cite{shade1998layered}, but in our case the pixels in each layer are fixed at a certain depth, and we use an alpha channel per layer to encode visibility. To render from \aPDIshort, the layers are composed from back-to-front order using the standard ``over'' alpha compositing operation. Figure~\ref{fig:representation} illustrates \aPDIshort. The \PDIshort representation is also related to the ``selection-plus-color'' layers used in DeepStereo~\cite{flynn2016deepstereo}, as well as to the volumetric representation of Penner and Zhang~\shortcite{penner2017soft3d}. 

We chose \PDIshorts because of their ability to represent geometry and texture including occluded elements, and because the use of alpha enables them to capture partially reflective or transparent objects as well as to deal with soft edges. Increasing the number of planes (which we can think of as increasing the resolution in disparity space) enables \aPDIshort to represent a wider range of depths and allows a greater degree of camera movement. Furthermore, rendering views from an MPI is highly efficient, and could allow for real-time applications.

\begin{figure*}[h]
\centering
\includegraphics[width=\linewidth]{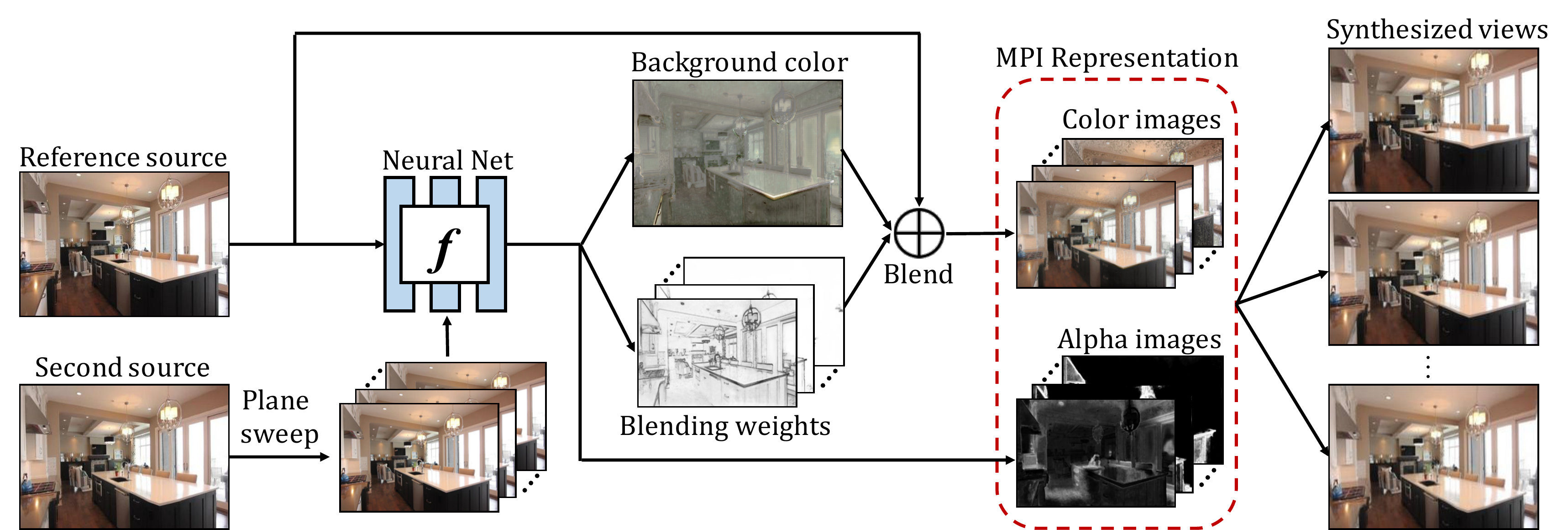}
\caption{Overview of our end-to-end learning pipeline. Given an input stereo image pair, we use a fully-convolutional deep network to infer the multiplane image representation. For each plane, the alpha image is directly predicted by the network, and the color image is blended by using the reference source and the predicted background image, where the blending weights are also output from the network. During training, the network is optimized to predict an MPI representation that reconstructs the target views using a differentiable rendering module (see Section~\ref{sec:viewsynth}). During testing, the MPI representation is only inferred once for each scene, which can then be used to synthesize novel views with minimal computation (homography + alpha compositing).}
\label{fig:pipeline}
\end{figure*}

Our representation recalls the \textit{multiplane camera} invented at Walt Disney Studios and used in traditional animation~\cite{wikimultiplanecamera}. In both systems, a scene is composed of a series of partially transparent layers at different distances from the camera.

\subsection{Learning from stereo pairs}\label{sec:learning}
We now describe our pipeline~(see Figure~\ref{fig:pipeline}) for learning a neural net that infers \PDIshorts from stereo pairs. In addition to the input images $I_1$ and $I_2$, we take as input their corresponding camera parameters $c_1 = (p_1, k_1)$ and $c_2 = (p_2, k_2)$, where $p_i$ and $k_i$ denote camera extrinsics (position and orientation) and intrinsics, respectively.

The reference coordinate frame for our predicted scene is placed at the camera center of the first input image $I_1$ (i.e., $p_1$ is fixed to be the identity pose). Our training set consists of a large set of $\langle I_1, I_2, I_t, c_1, c_2, c_t\rangle$ tuples, where $I_t$ and $c_t = (p_t, k_t)$ denote the target ground-truth image and its camera parameters, respectively. We aim to learn a neural network, denoted by $f_{\theta}(\cdot)$, that infers \aPDIshort representation using $\langle I_1, I_2, c_1, c_2\rangle$ as input, such that when the \PDIshort is rendered at $c_t$ it should reconstruct the target image $I_t$.

\paragraph{Network input}
To encode the pose information from the second input image $I_2$, we compute a plane sweep volume (PSV) that reprojects $I_2$ into the reference camera at a set of $D$ fixed depth planes.\footnote{For a rectified stereo pair, reprojected images would simply be shifted versions of $I_2$, though we consider more general configurations in our setup.} Although not required, we choose these depth planes to coincide with those of the output \PDIshort. This plane sweep computation results in a stack of reprojected images $\{\hat{I}_2^1, \ldots, \hat{I}_2^D \}$, which we concatenate along the color channels, resulting in a $H\times W \times 3D$ tensor $\hat{I}_2$. We further concatenate $\hat{I}_2$ with $I_1$ to obtain the input tensor (of size $H\times W\times 3(D+1)$) to the network. Intuitively, the PSV representation allows the network to reason about the scene geometry by simply comparing $I_1$ to each planar reprojection of $I_2$---the scene depth at any given pixel is typically at the depth plane where $I_1$ and the reprojected $I_2$ agree. Many stereo algorithms work on this principle, but here we let the network automatically learn such relationships through the view synthesis objective.

\paragraph{Network output}~A straightforward choice of the network output would be a separate RGBA image for each depth plane, where the color image captures the scene appearance and the alpha map encodes the visibility and transparency. However, such an output would be highly over-parameterized, and we found a more parsimonious output to be beneficial. In particular, we assume the color information in the scene can be well modeled by just two images, a foreground and a background image, where the foreground image is simply the reference source $I_1$, and the background image is predicted by the network, and is intended to capture the appearance of hidden surfaces. Hence, for each depth plane, we compute each RGB image $C_d$ as a per-pixel weighted average of the foreground image $I_1$ and the predicted background image $\hat{I}_b$:
\begin{equation}
    C_d = w_d \odot I_1 + (1 - w_d) \odot \hat{I}_b~,
\end{equation}
where $\odot$ denotes the Hadamard product, and the blending weights $w_d$ are also predicted by the network. Intuitively, $I_1$ would have a higher weight at nearer planes where foreground content is dominant, while $\hat{I}_b$ is designed to capture surfaces that are occluded in the reference view.
\new{
Note that the background image need not itself be a natural image, since the network can exploit the alpha and blending weights to selectively and softly use different parts of it at different depths. Indeed, there may be regions of a given background image that are never used in new views.
}

In summary, the network outputs the following quantities: 1) an alpha map $\alpha_d$ for each plane, 2) a global RGB background image $\hat{I}_b$ and 3) a blending weight image $w_d$ for each plane representing the relative proportion of the foreground and background layers at each pixel. If we predict $D$ depth layers each with a resolution of $W \times H$, then the total number of output parameters is $WH\cdot(2D + 3)$ (vs.\ $WH\cdot 4D$ for a direct prediction of \aPDIshort). These quantities can then be converted to \aPDIshort.

\subsection{Differentiable view synthesis using \PDIshorts}\label{sec:viewsynth}

Given the \PDIshort representation with respect to a reference frame, we can synthesize a novel view $\hat{I}_t$ by applying a planar transformation (inverse homography) to the RGBA image for each plane, followed by a alpha-composition of the transformed images into a single image in a back-to-front order. Both the planar transformation and alpha composition are differentiable, and can be easily incorporated into the rest of the learning pipeline.

\paragraph{Planar transformation}~Here we describe the planar transformation that inverse warps each \PDIshort RGBA plane
onto a target viewpoint. Let the geometry of the \PDIshort plane to be
transformed (i.e. the source) be $\mathbf{n}\cdot\mathbf{x} + a = 0$, where
$\mathbf{n}$ denotes the plane normal, $\mathbf{x} = [u_s, v_s, 1]^T$ the source
pixel homogeneous coordinates, and $a$ the plane offset. Since the source
\PDIshort plane is fronto-parallel to the reference source camera, we have
$\mathbf{n} = [0, 0, 1]$ and $a = -d_s$, where $d_s$ is the depth of the source
\PDIshort plane. The rigid 3D transformation matrix mapping from source to
target camera is defined by a 3D rotation $R$ and translation $\mathbf{t}$,
and the source and target camera intrinsics are denoted $k_s$ and $k_t$, respectively. Then for each pixel $(u_t, v_t)$ in the target \PDIshort plane, we use the standard inverse homography~\cite{hartley2003multiple} to obtain
\begin{gather}
    \begin{bmatrix}
    u_s \\ 
    v_s \\
    1
    \end{bmatrix} ~ \sim ~
    ~ k_s~ \left(R^{T} + \frac{R^T\mathbf{t}\mathbf{n}R^T}{a - \mathbf{n}R^T\mathbf{t}}\right) ~k_t^{-1}
    \begin{bmatrix}
    u_t \\ 
    v_t \\
    1
    \end{bmatrix}
\end{gather}
Therefore, we can obtain the color and alpha values for each target pixel $[u_t, v_t]$ by looking up its correspondence $[u_s, v_s]$ in the source image. Since $[u_s, v_s]$ may not be an exact pixel coordinate, we use bilinear interpolation among the 4-grid neighbors to obtain the resampled values (following~\cite{jaderberg2015spatial,zhou2016view}).

\paragraph{Alpha compositing}~After applying the planar transformation to each \PDIshort plane,
we then obtain the predicted target view by alpha compositing the color images
in back-to-front order using the standard \emph{over}
operation~\cite{porter1984compositing}.



\subsection{Objective}
Given the \PDIshort inference and rendering pipeline, we can train a network to
predict \PDIshorts satisfying our view synthesis objective. Formally, for a
training set of $\langle I_1, I_2, I_t, c_1, c_2, c_t\rangle$ tuples, we
optimize the network parameters by:
\begin{equation}
    \min_{\theta}~\sum_{\langle I_1, I_2, I_t, c_1, c_2, c_t\rangle}\mathcal{L}(\mathcal{R}(f_{\theta}(I_1, I_2, c_1, c_2), c_t), I_t)~,
\end{equation}
where $\mathcal{R}(\cdot)$ denotes the rendering pipeline described in Section~\ref{sec:viewsynth} that synthesizes a novel view from the target camera $c_t$ using the inferred \PDIshort $f_{\theta}(I_1, I_2, c_1, c_2)$, and $\mathcal{L}(\cdot)$ is the loss function between the synthesized view and the ground-truth. In this work, we use a deep feature matching loss (also referred to as the ``perceptual loss''~\cite{johnson2016perceptual,dosovitskiy2016generating,zhang2018perceptual}), and specifically use the normalized VGG-19~\cite{simonyan2014very} layer matching from~\cite{chen2017photographic}:
\begin{equation}
    \mathcal{L}(\hat{I}_t, I_t) = \sum_{l}\lambda_l \| \phi_l(\hat{I}_t) - \phi_l(I_t) \|_1~,
\end{equation}
where $\{ \phi_l \}$ is a set of layers in VGG-19 (\texttt{conv1\_2}, \texttt{conv2\_2}, \texttt{conv3\_2}, \texttt{conv4\_2}, and \texttt{conv5\_2}) and the weight hyperparameters $\{\lambda_l \}$ are set to the inverse of the number of neurons in each layer.

\subsection{Implementation details}\label{sec:impl}
Unless specified otherwise, we use $D=32$ planes set at equidistant disparity (inverse depth) with the near and far planes at $1\mathrm{m}$ and $100\mathrm{m}$, respectively.

\paragraph{Network architecture}~We use a fully-convolutional encoder-decoder architecture (see Table~\ref{tab:net} for detailed specification). The encoder pathway follows similar design as VGG-19 \cite{simonyan2014very}, while the decoder consists of deconvolution (fractionally-strided convolution) layers with skip-connections from lower layers to capture fine texture details. Dilated convolutions~\cite{yu2015multi,chen2018deeplab} are also used in intermediate layers \texttt{conv4\_1,2,3} to model larger scene context while maintaining the spatial resolution of the feature maps. Each layer is followed by a \texttt{ReLU} nonlinearity and layer normalization~\cite{ba2016layer} except for the last layer, where \texttt{tanh} is used and no layer normalization is applied. Each of the last layer outputs (32 alpha images, 32 blending weight images, and 1 background RGB image) is further scaled to match the corresponding valid range (e.g. $[0, 1]$ for alpha images).

\paragraph{Training details}~We implement our system in TensorFlow~\cite{abadi2016tensorflow}. We train the network using the ADAM solver \cite{kingma2014adam} for 600K iterations with learning rate $0.0002, \beta_1 = 0.9, \beta_2 = 0.999,$ and batch size $1$. During training, the images and \PDIshort have a spatial resolution of $1024\times 576$, but the model can be applied to arbitrary resolution at test time in a fully-convolutional manner. Training takes about one week on a Tesla P100 GPU.

\begin{table}[t]
\centering{
\caption{Our network architecture, where \textbf{k} is the kernel size, \textbf{s} the stride, \textbf{d} kernel dilation, \textbf{chns} the number of input and output channels for each
layer, \textbf{in} and \textbf{out} are the accumulated stride for the input and output of each layer, and \textbf{input} denotes the input source of each layer with $+$ meaning concatenation. See Section~\ref{sec:impl} for more details.}
\label{tab:net}
\begin{tabular}{cccccccc}
\toprule
  \textbf{Layer} & \textbf{k} & \textbf{s} & \textbf{d} & \textbf{chns} & \textbf{in} & \textbf{out} & \textbf{input}
\tabularnewline
\midrule
conv1\_1 & 3 & 1 & 1 & 99/64   & 1 & 1 & $I_1 + \hat{I}_2$
\tabularnewline
conv1\_2 & 3 & 2 & 1 & 64/128  & 1 & 2 & conv1\_1
\tabularnewline
conv2\_1 & 3 & 1 & 1 & 128/128 & 2 & 2 & conv1\_2
\tabularnewline
conv2\_2 & 3 & 2 & 1 & 128/256 & 2 & 4 & conv2\_1
\tabularnewline
conv3\_1 & 3 & 1 & 1 & 256/256 & 4 & 4 & conv2\_2
\tabularnewline
conv3\_2 & 3 & 1 & 1 & 256/256 & 4 & 4 & conv3\_1
\tabularnewline
conv3\_3 & 3 & 2 & 1 & 256/512 & 4 & 8 & conv3\_2
\tabularnewline
conv4\_1 & 3 & 1 & 2 & 512/512 & 8 & 8 & conv3\_3
\tabularnewline
conv4\_2 & 3 & 1 & 2 & 512/512 & 8 & 8 & conv4\_1
\tabularnewline
conv4\_3 & 3 & 1 & 2 & 512/512 & 8 & 8 & conv4\_2
\tabularnewline
\midrule
conv5\_1 & 4 & .5 & 1 & 1024/256 & 8 & 4 & conv4\_3 + conv3\_3
\tabularnewline
conv5\_2 & 3 &  1 & 1 &  256/256 & 4 & 4 & conv5\_1
\tabularnewline
conv5\_3 & 3 &  1 & 1 &  256/256 & 4 & 4 & conv5\_2
\tabularnewline
conv6\_1 & 4 & .5 & 1 &  512/128 & 4 & 2 & conv5\_3 + conv2\_2
\tabularnewline
conv6\_2 & 3 &  1 & 1 &  128/128 & 2 & 2 & conv6\_1
\tabularnewline
conv7\_1 & 4 & .5 & 1 &   256/64 & 2 & 1 & conv6\_2 + conv1\_2
\tabularnewline
conv7\_2 & 3 &  1 & 1 &   64/64  & 1 & 1 & conv7\_1
\tabularnewline
conv7\_3 & 1 &  1 & 1 &   64/67  & 1 & 1 & conv7\_2
\tabularnewline
\bottomrule
\end{tabular}}
\end{table}

\section{Data}

For training we require triplets of images together with their relative camera poses and intrinsics. Creating such a dataset from scratch would require carefully capturing simultaneous photos of a variety of scenes from three or more appropriate viewpoints per scene. Instead, we identified an existing source of massive amounts of such data: video clips on YouTube shot from a moving camera. By sampling frames from such videos, we can obtain very large amounts of data comprising multiple views of the same scene shot from a variety of baselines. 
For this approach to work, we need to be able to identify suitable video clips, i.e., clips shot from a moving camera but with a static scene, with minimal artifacts such as motion blur or rolling-shutter distortion, and without other editing effects such as titles and overlays. Finally, given a suitable clip, we must estimate the camera parameters for each frame.

While many videos on YouTube are not useful for our purposes, we found a surprisingly large amount of suitable content, across several categories of video. One such category is real estate footage. 
Typical real estate videos feature a series of shots of indoor and outdoor scenes (the interior of a room or stairway, exterior views of a house, footage of the surrounding area, etc). Shots typically feature smooth camera movement and little or no scene movement. 
Hence, we decided to build a dataset from real estate videos as a large and diverse source of multi-view training imagery.

Accordingly, the rest of this section describes the dataset we collected, consisting of over 7,000 video clips from 1 to 10 seconds in length, together with the camera position, orientation and field of view for each frame in the sequence. To build this dataset, we devised a pipeline for mining suitable clips from YouTube. This pipeline consists of four main steps: 1) identifying a set of candidate videos to download, 2) running a camera tracker on each video to both estimate an initial camera pose for each frame and to subdivide the video into distinct shots/clips, 3) performing a full bundle adjustment to derive high-quality poses for each clip, and 4) filtering to remove any remaining unsuitable clips.

\subsection{Identifying videos}
We manually found a number of
YouTube channels that published real estate videos exclusively or almost exclusively, and used the YouTube API 
to retrieve videos IDs listed under each channel. This yielded a set of approximately 1,500 candidate videos.

\subsection{Identifying and tracking clips with SLAM}
We wish to subdivide each video into individual clips, and identify clips that have significant camera motion. We found few readily available tools for performing camera tracking on arbitrary videos in the wild. Initially, we tried to use structure-from-motion methods developed in computer vision, such as \textsc{Colmap}~\cite{schoenberger2016sfm}. These methods are optimized for photo collections, and we found them to be slow and prone to failure when applied to video sequences.
Instead, we found that for our purposes we could adapt modern algorithms for SLAM (Simultaneous Localization and Mapping) developed in the robotics community.

Visual SLAM methods take as input a series of frames, and build and maintain a sparse or semi-dense 3D reconstruction of the scene while estimating the viewpoint of the current frame in a way consistent with this reconstruction. We use the \OrbSlam system~\cite{murartal2015orbslam}, though other methods could also apply~\cite{forster2014svo,engel2018direct}.

SLAM algorithms are not designed to process videos containing multiple shots
with cuts and dissolves between them, and they typically care only about the
accuracy of the \textit{current} frame's pose---in particular, as the scene is
refined over time, earlier frames are not updated and may become inconsistent
with the current state of the world. To deal with these issues, our approach is
as follows: \textbf{1}. Feed successive frames of the video to \OrbSlam as
normal. \textbf{2}. When the algorithm reports that it has begun to track the
camera, mark the start of a clip. \textbf{3}. When \OrbSlam fails to track $K=6$
consecutive frames, or when we reach a maximum sequence length $L$, consider the
clip to have ended. \textbf{4}. Keeping the final scene model constant,
reprocess all frames in the clip so as to estimate a consistent pose for each
camera. \textbf{5}. Re-initialize \OrbSlam so it is ready to start tracking a
new clip on subsequent frames. In this way, we use \OrbSlam not just to track
frames, but also to divide a video into clips using tracking failure as a way to
detect shot boundaries.

Since SLAM methods, including \OrbSlam, require known camera intrinsics such as
field of view (which are unknown for arbitrary online videos), we simply assume
a field of view of 90 degrees. This assumption worked surprisingly well for the
purposes of identifying good clips. Finally, for the sake of speed, at this
stage we process a lower resolution version of the video. The result of the
above processing is a set of clips or sequences for each video, along with a
preliminary set of camera parameters.

\subsection{Refining poses with bundle adjustment}

\begin{figure}[t]
\begin{center}
\includegraphics[width=\linewidth]{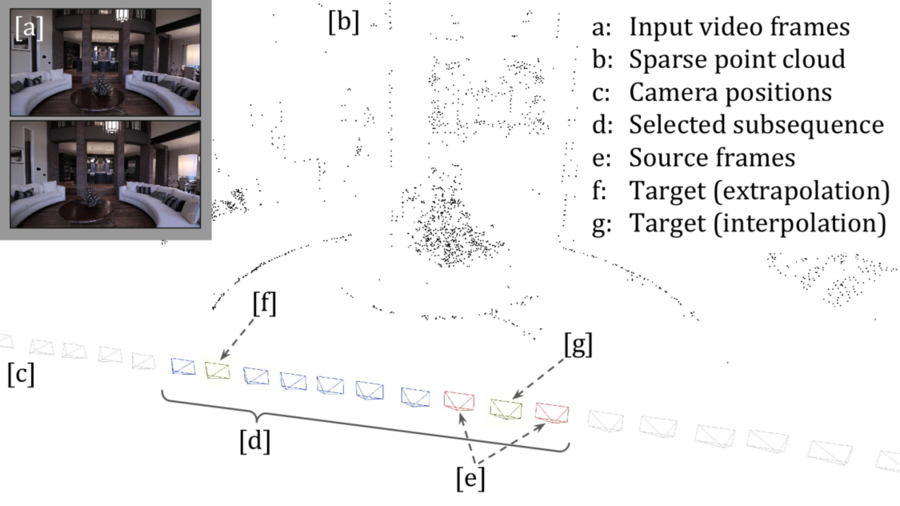}
\caption{Dataset output and frame selection, showing estimated camera trajectory and sparse point cloud. See section~\ref{sec:triples} for a detailed description.}
\label{fig:data}
\end{center}
\end{figure}

We next process each sequence at higher resolution, using a standard structure-from-motion pipeline to extract features from each frame, match these features across frames, and perform a global bundle adjustment using the Ceres non-linear least squares optimizer~\cite{ceres-solver}. We initialize the cameras using the poses found by \OrbSlam, and add a weak penalty to the optimization that encourages the parameters not to stray too far from their initial values. The output for each sequence is a set of adjusted camera poses, an estimated field of view, and a sparse point cloud representing the scene. An example output is illustrated in Figure~\ref{fig:data}.

One difficulty with this process is that there is no way to determine global scene scale, so our reconstructed camera poses are up to an arbitrary scale per clip. This ambiguity will become important when we represent scenes with \PDIshorts, because our representation is based on layers at specific depths, as described in Section~\ref{sec:impl}. Hence, we ``scale-normalize'' each sequence using the estimated 3D point cloud, scaling it so that the nearest scene geometry is approximately a fixed distance from the cameras. In particular, for each frame we compute the 5th percentile depth among all point depths from that frame's camera. Computing this depth across all cameras in a sequence gives us a set of ``near plane'' depths. We scale the sequence so that the 10th percentile of this set of depths is 1.25m. (Recall that our \PDIshort representation uses a near plane of 1m.)

\subsection{Filtering and clipping}

If the source video contains cross-fades, some frames may show a blend of two scenes. We discard ten frames from the beginning and end of each clip, which eliminates most such frames.

Occasionally the estimated camera poses for a sequence do not form a smooth track, which can indicate that we were unable to track the camera accurately. We define a frame to be \textit{smooth} if its camera position $p_i$ is sufficiently close to the average of the two adjacent camera positions, specifically if $\|p_i - (p_{i+1} + p_{i-1})/2\| < 0.2\times\|p_{i+1} - p_{i-1}\|$. For each sequence, we find the longest consecutive subsequence in which all frames are smooth, and discard the rest.

Finally we discard all remaining sequences of fewer than 30 frames. From an input set of approximately 1500 videos, this pipeline produces a set of $\sim$7,000 sequences with a total of $\sim$750K frames.

\subsection{Choosing training triplets}\label{sec:triples}

Figure~\ref{fig:data} shows an example of the result of this processing, including input video frames [a] (just two frames are shown here), and the sparse point cloud [b] and camera track [c] resulting from the structure from motion pipeline. As described in Section~\ref{sec:learning}, for our application we require tuples $\langle I_1, I_2, I_t, c_1, c_2, c_t\rangle$, including cases where $I_t$ is an \emph{extrapolation} from $I_1$ and $I_2$. We sample tuples from our dataset by first selecting from each sequence a random subsequence [d] of length 10, with stride (gap between selected frames) chosen randomly from 1 to 10. From this subsequence we then randomly choose two different frames and their poses to be the inputs $I_1, I_2, c_1,$ and $c_2$ [e], and a third frame to be the target $I_t, c_t$.

Depending on which frames are chosen, the target frame may require extrapolation [f] (of up to a factor of nine times the distance between $I_1$ and $I_2$, assuming a linearly moving camera) or interpolation [g] from the inputs. We chose to learn to predict views from a variety of positions relative to the source imagery so as not to overfit to generating images at a particular distance during training.

\section{Experiments and results}
In this section we evaluate the performance of our method, and compare it with several view synthesis baselines. Our test set consists of 1,329 sequences that did not overlap with the training set. For each sequence we randomly sample a triplet (two source frames and one target frame) for evaluation. We first visualize the MPI representation inferred by our model, and then provide detailed comparison with other recent view synthesis methods. We further validate our model design with various ablation studies, and finally highlight the utility of our method through several applications. For quantitative evaluation, we use the standard SSIM~\cite{wang2004image} and PSNR metrics.

\subsection{Visualizing the \pdis}
\begin{figure*}[t]
\centering
\includegraphics[width=\linewidth]{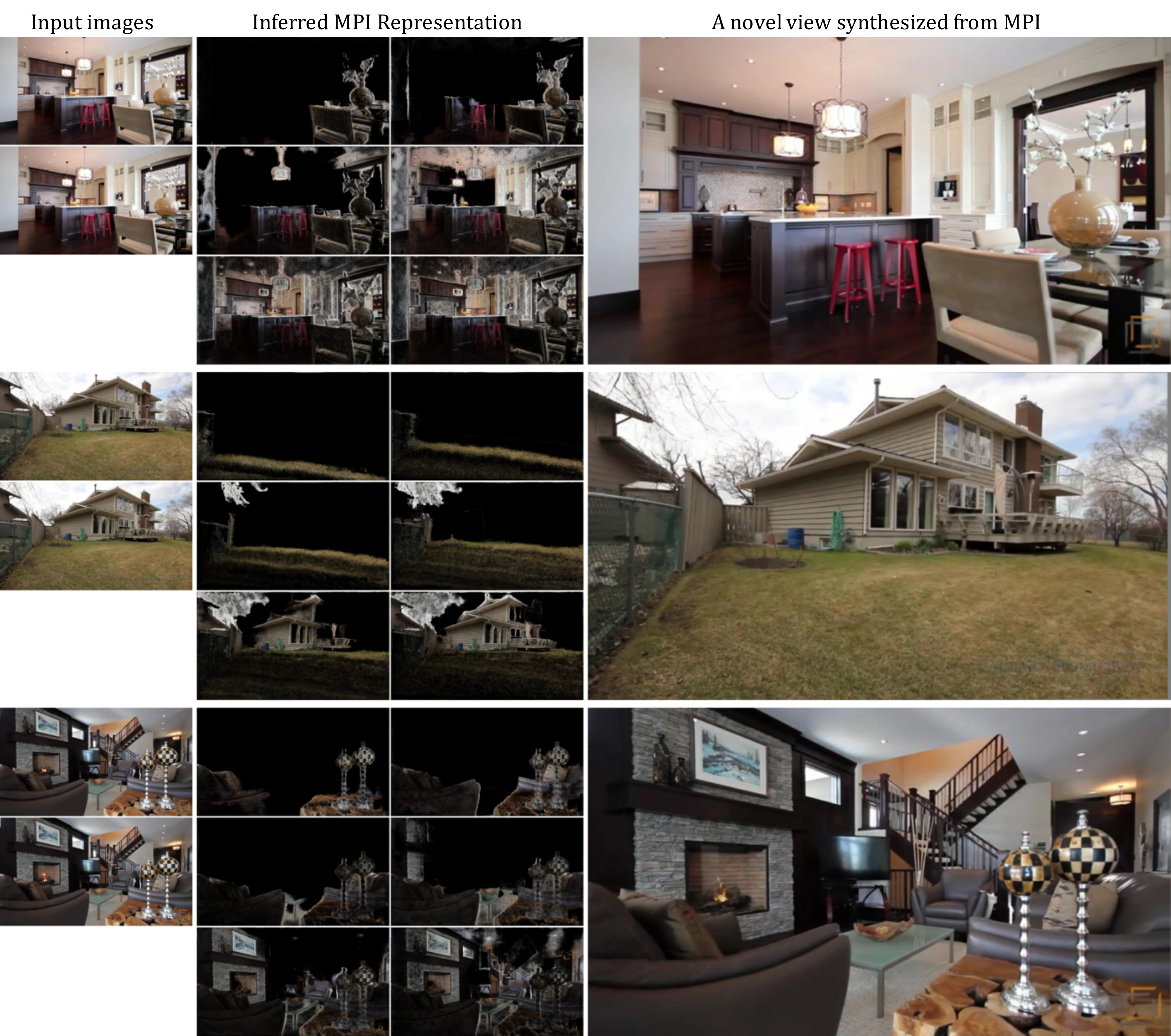}
\caption{Sample visualization of the input image pair (left), our inferred \PDIshort representation (middle), where we show the alpha-multiplied color image at a subset of the depth planes from near to far (top to bottom, left to right), and novel views rendered from the \PDIshort (right). The predicted \PDIshort is able to capture the scene appearance in a layer-wise manner (near to far) respecting the scene geometry. }
\label{fig:results}
\end{figure*}

We visualize examples of the \PDIshort representation inferred by our network in Figure~\ref{fig:results}. Despite having no direct color or alpha ground-truth for each \PDIshort plane during training, the inferred \PDIshort is able to capture the scene appearance in a layer-wise manner (near to far) respecting the scene geometry, which allows realistic rendering of novel views from the representation.

\begin{figure*}[t]
\centering
\includegraphics[width=\linewidth]{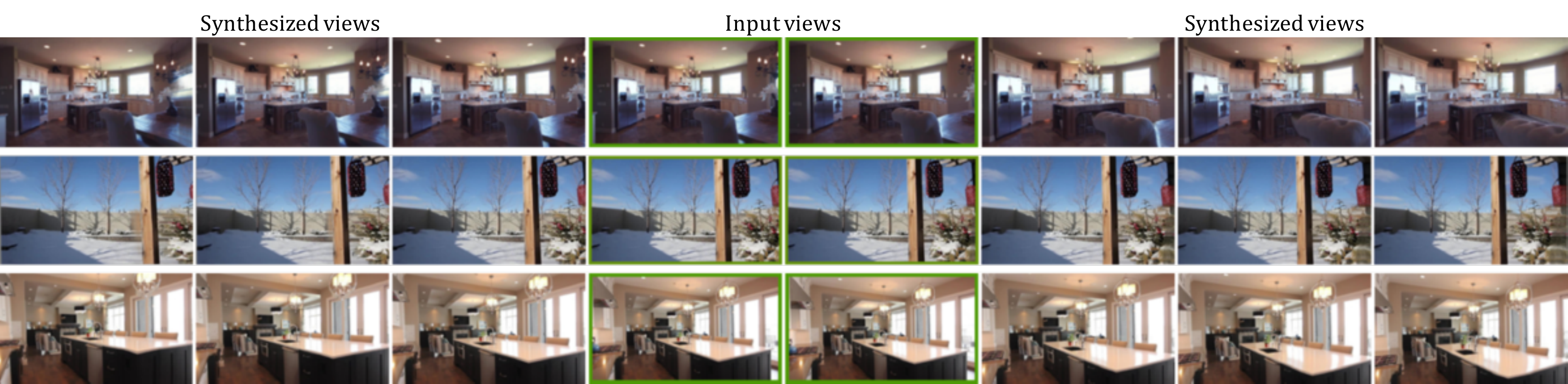}
\caption{Sample view extrapolation results using \pdis. The central two frames (green) are the input to our network, and the inferred \PDIshort is used to render both past and future frames in the same video sequence.}
\label{fig:extrap}
\end{figure*}

We also demonstrate view extrapolation capability of the \PDIshort representation in Figure~\ref{fig:extrap}, where we use the central two frames of a registered video sequence as input, and synthesize the previous and future frames with the inferred \PDIshort. Please see the supplemental video for animations of these rendered sequences.

\subsection{Comparison with Kalantari et al.}\label{sec:kala}
We compare our model with Kalantari et al.~\shortcite{kalantari2016learning}, a state-of-the-art learning-based view synthesis method. A critical difference compared to our method is that Kalantari et al. has an \emph{independent} rendering process for each novel view of the scene, and needs to re-run the entire inference pipeline every time a new view is queried, which is computationally prohibitive for real-time applications. In contrast, our method predicts a scene-level \PDIshort representation that can render any novel viewpoint in real-time with minimal computation (inverse homography + alpha compositing).

We train and test \new{two} variants of their method on our data: 1) same network architecture (4 convolution layers) and pixel reconstruction loss from the original paper; 2) our network architecture (which is deeper with skip connections) with perceptual loss. For fair comparison, we use the same number of input planes as ours for constructing the plane sweep volume in their input. \new{See Section~\ref{sec:ablation} for discussion on the effect of varying the number of depth planes.}

\begin{table}[t]
\centering{
\caption{Quantitative comparison between our model and variants of the baseline Kalantari model~\shortcite{kalantari2016learning}. \new{Higher SSIM/PSNR mean and lower rank are better}. See Section~\ref{sec:kala} for more details.}
\label{tab:kala}
\new{\begin{tabular}{lcc|cccc}
\toprule
 Method & Network & Loss & \multicolumn{2}{c}{SSIM} & \multicolumn{2}{c}{PSNR} \tabularnewline
 & & & Mean & Rank & Mean & Rank \tabularnewline
\midrule
Kalantari & Kalantari & pixel  & $0.696$ & $4.0$ & $31.41$ & $3.7$  \tabularnewline
Kalantari & Ours & VGG         & $0.822$ & $2.1$ & $32.93$ & $2.0$  \tabularnewline
Ours & Ours & Pixel            & $0.812$ & $2.6$ & $32.42$ & $2.8$ \tabularnewline
\textbf{Ours} & \textbf{Ours} & \textbf{VGG}  & $\mathbf{0.835}$ & $\mathbf{1.4}$ & $\mathbf{33.10}$ & $\mathbf{1.5}$ \tabularnewline
\bottomrule
\end{tabular}}}

\end{table}

\new{Table~\ref{tab:kala} shows mean SSIM and PSNR similarity metrics for each method across our test set. To measure if one method is consistently better than another, we also rank the methods on each test triplet and compute the average rank for each method. An average rank of 1.0 for PSNR, for example, would mean that this method always had the highest PSNR score.}

We find that 1) our network architecture is significantly more effective than the simple 4-layer network used in the original Kalantari paper; 2) the VGG perceptual loss helps improve the performance over the pixel reconstruction loss (see Section~\ref{sec:ablation} for discussion); \new{3) our model outperforms the better of the two Kalantari variants} (VGG with our network architecture), indicating the high-quality of novel views rendered from the \PDIshort representation.

We also observe that when rendering continuous view sequences of the same scene, our results tend to be more spatially coherent than Kalantari, and produce fewer frame-to-frame artifacts. We hypothesize that this is because, unlike the Kalantari model, we infer a single scene-level \PDIshort representation that is shared for rendering all target views, which implicitly imposes a smoothness prior when rendering nearby views. Please see the video for qualitative comparisons of our method to Kalantari on rendered sequences.

\begin{figure}[t]
\centering
\includegraphics[width=\linewidth]{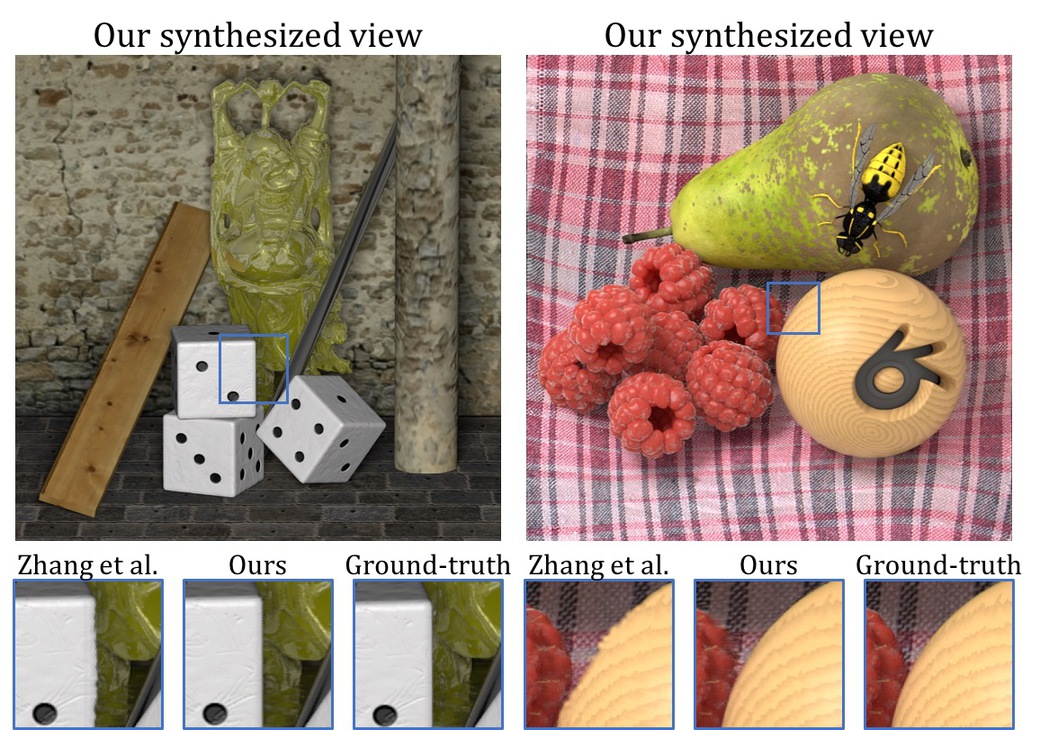}
\caption{\new{Comparison with Zhang et al.~\shortcite{zhang2015light} on the HCI light field dataset~\cite{wanner2013datasets}. Note the differences around object boundaries.}}
\label{fig:zhang}
\end{figure}

\new{
\subsection{Comparison with extrapolation methods}\label{sec:zhang}
We compare with a non-learning view extrapolation approach by Zhang et al.~\shortcite{zhang2015light}, which reconstructs a 4D light field from micro-baseline stereo pairs using disparity-assisted phase based synthesis (DAPS). For fair comparison, we directly apply our model trained on the real estate data to the HCI light field dataset~\cite{wanner2013datasets}. As shown in Figure~\ref{fig:zhang}, our model generalizes well on the HCI dataset without any fine-tuning, and compares favorably with Zhang et al.\ around depth boundaries, where our method introduces fewer distortion artifacts. We find that the method of Zhang et al.\ performs well for small view extrapolations, but breaks down more quickly around object boundaries  with increasing extrapolation distance.
}

\new{
We also trained appearance flow~\cite{zhou2016view} on our dataset, but found rendered views exhibited significant artifacts, such as straight lines becoming distorted. This method appears more suited to object-centric synthesis than to scene rendering, and it is not able to fully exploit correlations between views since the trained network operates on each input image separately.
}

\subsection{Ablation studies}\label{sec:ablation}

\begin{figure}[t]
    \centering
    \includegraphics[width=\linewidth]{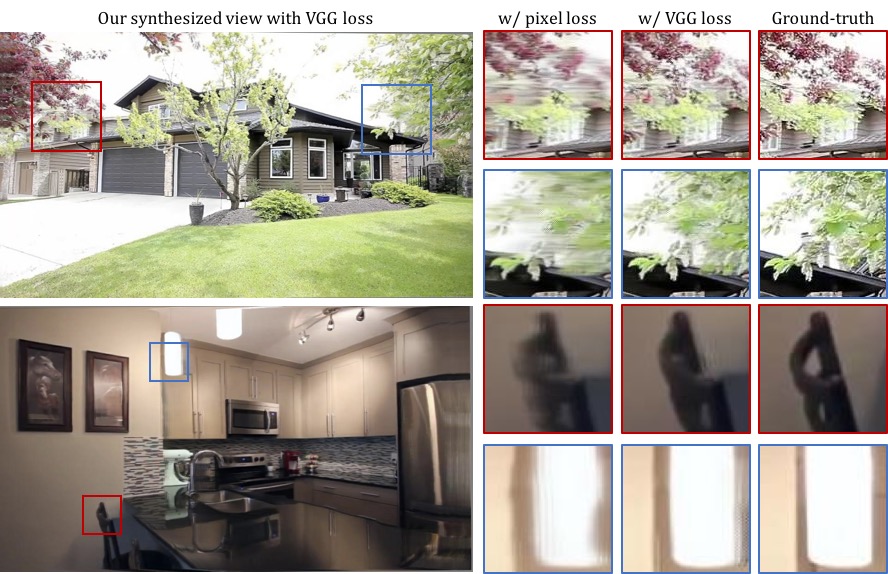}
    \caption{Comparison between the models trained using pixel reconstruction loss and VGG perceptual loss. The latter better preserves object structure, and tends to produce sharper synthesized views.}
    \label{fig:vgg}
\end{figure}

\paragraph{Perceptual loss}~To illustrate the effect of the perceptual loss, we compare our final model with a baseline model trained using L1 loss in the RGB pixel space. As shown in Figure~\ref{fig:vgg}, our final model trained using the perceptual loss better preserves object structure and texture details in the synthesized results than the baseline. The benefit of training with perceptual loss is further verified with quantitative evaluation in Table~\ref{tab:kala}.

\begin{figure}[t!]
    \centering
    \includegraphics[width=\linewidth]{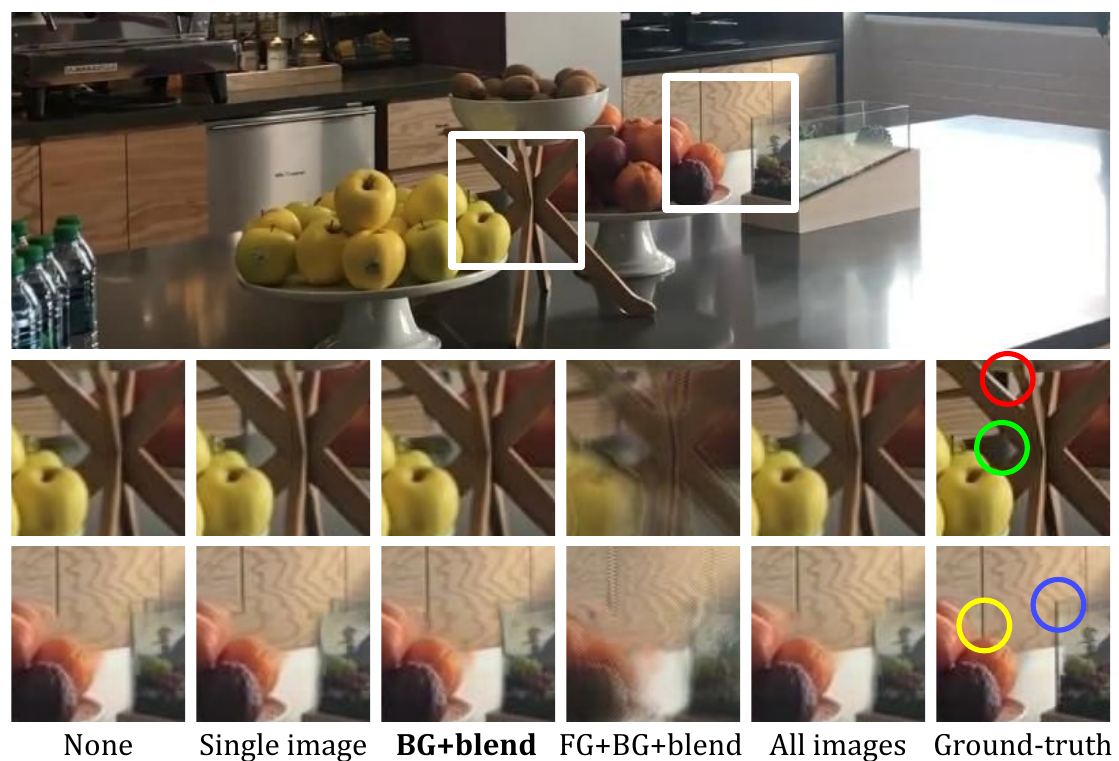}
    \caption{\new{Comparison between different color prediction formats. Note in particular the rendering of disoccluded background details, such as the rear wall (red), its reflection in the table surface (green), cupboard door (yellow) and corner of vase (blue). All the variants (except ``FG+BG+blend") produce competitive results with slight differences. See Section~\ref{sec:ablation} for more details.}}
    \label{fig:color}
\end{figure}

\paragraph{Color layer prediction}~In Section~\ref{sec:learning}, we propose that our network create the color values for each \PDIshort plane as a weighted average of a network predicted ``background'' image and the reference source image. Here we compare several variants of the color prediction format (ordered by increasing level of representation flexibility):
\begin{enumerate}
  \item None. No color image or blending weights are predicted by the network. The reference source image is used as the color image at each \PDIshort plane.
  \item Single image. The network predicts a single color image shared for all \PDIshort planes.
  \item Background + blending weights (our preferred format). The network predicts a background image and blending weights. The reference source is used as the foreground image.
  \item Foreground + background + blending weights. In contrast to the previous variant, instead of using the reference source as the foreground image, the network predicts an extra foreground image for blending with the background. 
  \item All images. The network directly outputs the color image at each \PDIshort plane.
\end{enumerate}
\new{
  We compare the performance of these variants in Table~\ref{tab:color} and show a qualitative example in Figure~\ref{fig:color}.
Although ``BG+blending weights'' slightly outperforms the other variants, all the variants (other than ``FG+BG+blending weights") produce competitive results. The ``None'' and ``Single image'' variants suffer in areas where the target view contains details that are occluded in the reference image but visible in the second input image. The ``BG+blending weights'' format can represent these areas better since not all \PDIshort planes need to have the same color data. The ``FG+BG+blending weights'' variant is slightly more powerful as the foreground image is not restricted, and the ``All images'' variant, with a separate color image for each plane, is the only variant that can fully represent a scene with depth complexity greater than 2. However, in our experiments these last two variants both performed slightly worse than ``None''. We hypothesize that the larger output space and less utilization of the reference image makes the learning harder with these output formats, and that the relatively small camera movement limits the depth complexity required.
}

\begin{table}[t]
\centering{\new{
\caption{Quantitative evaluation of variants of network color output, ordered by increasing degree of flexibility (top to bottom). \new{Higher SSIM/PSNR mean and lower rank are better.}}
\label{tab:color}
\begin{tabular}{l|cccc}
\toprule
Color layer prediction & \multicolumn{2}{c}{SSIM} & \multicolumn{2}{c}{PSNR} \tabularnewline
 & Mean & Rank & Mean & Rank \tabularnewline
\midrule
 None                        & $0.833$ & $2.3$ & $33.06$ & $2.1$ \tabularnewline
 Single image                & $0.822$ & $3.9$ & $32.51$ & $3.9$ \tabularnewline
 \textbf{BG + Blend weights} & $\mathbf{0.835}$ & $\mathbf{1.6}$ & $\mathbf{33.09}$ & $\mathbf{1.6}$ \tabularnewline
 FG + BG + Blend weights     & $0.819$ & $4.1$ & $32.50$ & $3.7$ \tabularnewline
 All images                  & $0.825$ & $3.2$ & $32.53$ & $3.8$ \tabularnewline
\bottomrule
\end{tabular}}}
\end{table}

\begin{figure}[t]
    \centering
    \includegraphics[width=\linewidth]{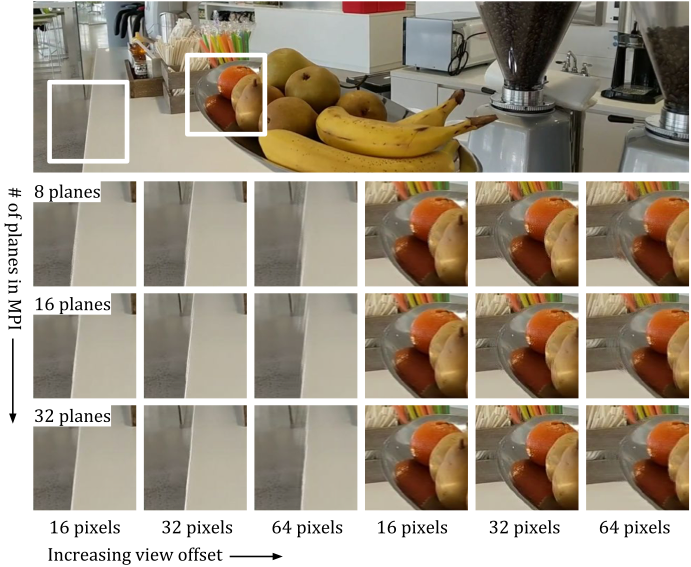}
    \caption{\new{Effect of varying the number of depth planes at different view offsets. For two regions of the top image, we show view extrapolations from \PDIshorts with varying numbers of planes. The number of pixels shown is the disparity between front and back planes relative to the reference view. The larger the number of planes, the farther the view can be extrapolated before introducing artifacts. Note the edge of the counter in the first example, and the edges of objects in the second example. (Best viewed zoomed in.)}}
    \label{fig:planes}
\end{figure}

\paragraph{Number of depth planes}~As shown in Table~\ref{tab:planes}, our model performance improves as more depth planes are used in the inferred \PDIshort representation. We are currently limited to 32 planes due to memory constraints, but could overcome this with future hardware or alternative networks.
\new{
As seen in Figure~\ref{fig:planes}, the greater the offset between the reference view and the rendered view, the more planes are needed to render the scene accurately.
}

\begin{table}[t]
\centering{
\caption{Evaluating the effect of varying the number of depth planes for the \PDIshort representation. \new{Higher SSIM/PSNR mean and lower rank are better.}}
\label{tab:planes}
\new{
\begin{tabular}{l|cccc}
\toprule
 \PDIshort depth planes & \multicolumn{2}{c}{SSIM} & \multicolumn{2}{c}{PSNR} \tabularnewline
 & Mean & Rank & Mean & Rank \tabularnewline
\midrule
D = 8           & $0.766$ & $2.99$ & $32.12$ & $2.96$ \tabularnewline
D = 16          & $0.812$ & $1.98$ & $32.73$ & $1.97$ \tabularnewline
\textbf{D = 32} & $\mathbf{0.835}$ & $\mathbf{1.03}$ & $\mathbf{33.09}$ & $\mathbf{1.07}$ \tabularnewline
\bottomrule
\end{tabular}}}
\end{table}

\subsection{Applications}

\begin{figure}[t]
    \centering
    \begin{tabular}{cc}
    \includegraphics[height=1.2in]{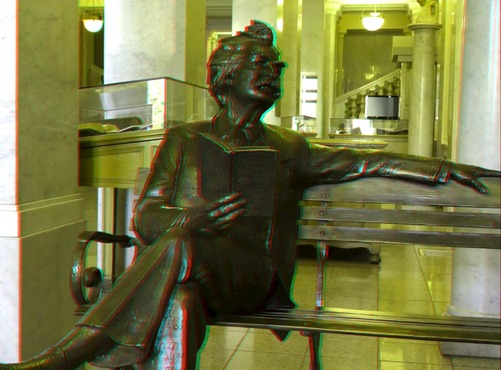} &
    \includegraphics[height=1.2in]{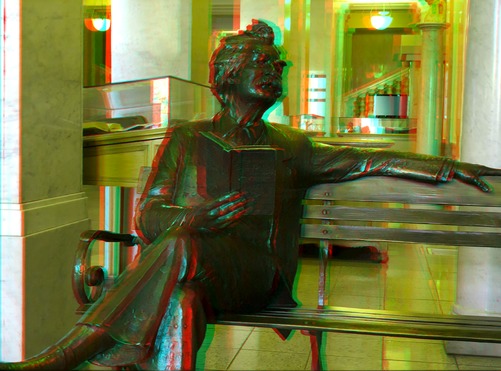}\\
    \includegraphics[height=1.2in]{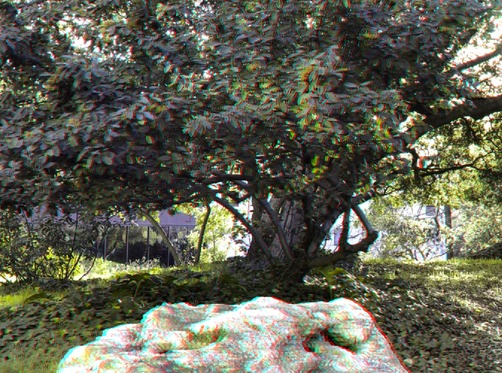} &
    \includegraphics[height=1.2in]{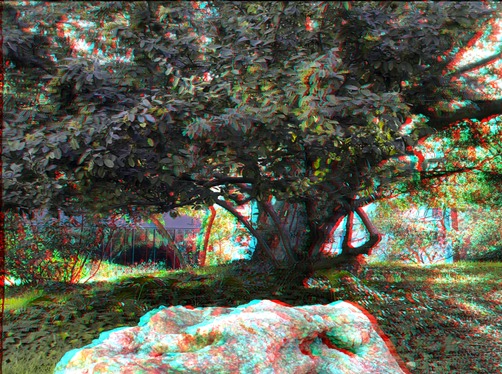}\\
    Original baseline & Magnified baseline
    \end{tabular}
    \caption{Example stereo magnifications for dual-lens camera. Left: raw stereo pairs captured by an iPhone X, displayed as red-cyan anaglyph images, with a baseline of $\sim$1.4cm. Right: the same images but with baseline synthetically magnified to $\sim$6.3cm. Note the significantly enhanced stereo effect. (Best viewed zoomed in and with 3D glasses.)}
    \label{fig:anaglyph}
\end{figure}

In this section we describe two applications of our trained model: 1) taking a narrow-baseline stereo pair from a cell phone camera and extrapolating to an average human interpupillary-distance (IPD)-spaced stereo pair, and 2) taking an image pair from a large-baseline stereo camera and extrapolating a ``1D lightfield'' of views between and beyond the source images. 

\paragraph{Cell phone image pairs $\rightarrow$ IPD stereo pair} We captured a set of image pairs with an iPhone X, a recent dual-lens camera phone with a baseline of $\sim$1.4cm, using an app that saves both captured views. Because the focal lengths of the two cameras are different, the app crops the wider-angle image to match the narrower field-of-view image. For each image pair, we ran a calibration procedure to refine the camera intrinsics
using their nominal values as initialization. We then applied our model (trained on real estate data) to magnify the baseline to $\sim$6.3cm (a magnification factor of 4.5x). Several results are shown in Figure~\ref{fig:anaglyph} as anaglyph images, and in the supplemental video as sway animation. Figure~\ref{fig:anaglyph} highlights how the extrapolated images provide a more compelling sense of 3D, and illustrates how our model can generalize to new scenarios that are atypical of real estate scenes (such as the sculpture of Mark Twain in the first example). Finally, notice that our method can handle interesting materials (e.g. the reflective glass and glossy floor in the first scene).

\begin{figure}[t]
    \centering
    \begin{tabular}{cc}
    \includegraphics[height=1.2in]{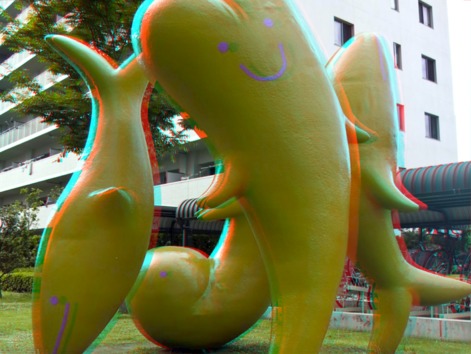} &
    \includegraphics[height=1.2in]{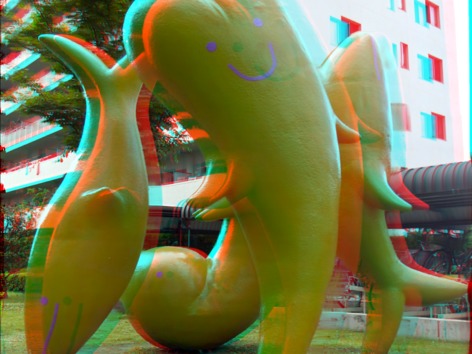}\\
    Original baseline & Magnified baseline
    \end{tabular}
    \caption{Example stereo magnifications for Fujifilm Real 3D stereo camera. Left: a raw stereo pair from the camera, displayed as red-cyan anaglyph images, with a baseline of $\sim$7.7cm. Right: the same images but with baseline synthetically magnified to $\sim$26.7cm. (Best viewed zoomed in and with 3D glasses.) (Photo used under CC license from Flickr user heiwa4126.)}
    \label{fig:pairs}
\end{figure}

\paragraph{Stereo pairs to extended 1D lightfield} We also demonstrate taking a large-baseline stereo pair and synthesizing a continuous ``1D lightfield''---i.e., a set of views along a line passing through the source views. For this application, we downloaded stereo pairs shot by a Fujifilm FinePix Real 3D W1 stereo point-and-shoot camera with a baseline of 7.7cm, and extrapolated to a continuous set of views with a baseline of 26.7cm (a magnification factor of $\sim$3.5x). Figure~\ref{fig:pairs} shows an example input and output as anaglyphs; see the supplemental video for animations of the resulting sequences. This input baseline, magnification factor, and scene content represent a challenging case for our model, and artifacts such as stretching in the background can be observed. Nonetheless, the results show plausible interpolations and extrapolations of the source imagery.

\begin{figure*}[t]
\centering
\includegraphics[width=\linewidth]{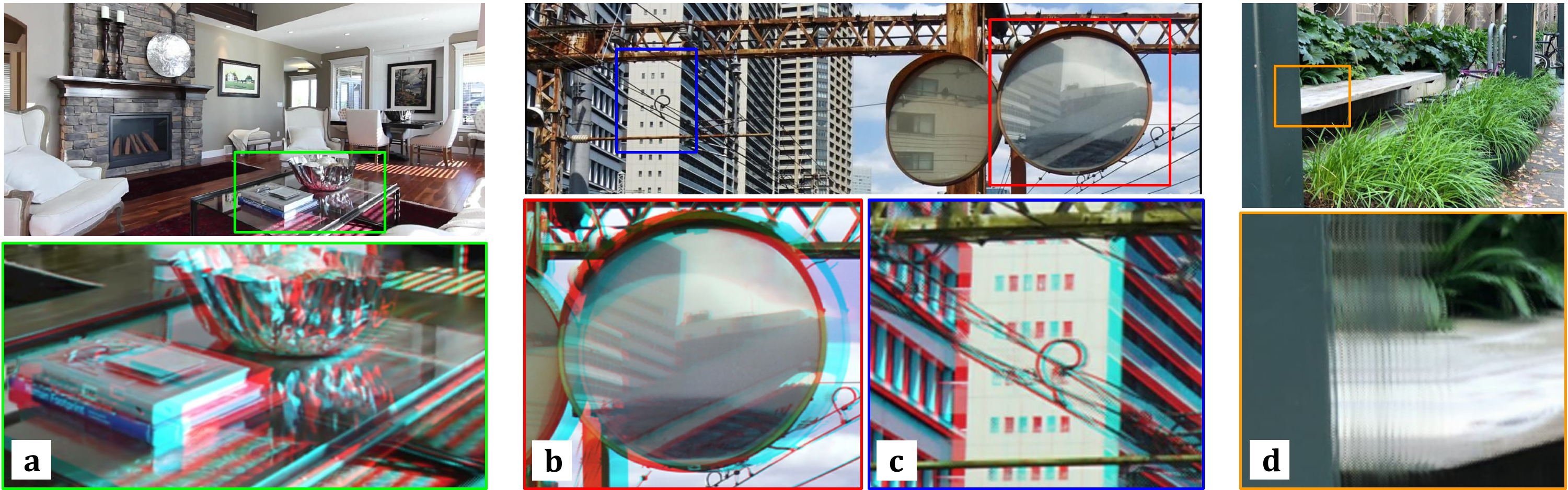}
\caption{Challenging cases. Reference images at top, rendered anaglyph details at bottom: (a) glass table with reflection and transparency, (b) reflection in a dusty curved mirror, (c) fine wires are confused with background, (d) extrapolation beyond the limits of the representation gives a `stack of cards' effect.}
\label{fig:discussion}
\end{figure*}

\section{Discussion}

Having trained on a large and varied dataset, our view synthesis system based on multiplane images is able to handle both indoor and outdoor scenes. We successfully applied it to scenes which are quite different from those in our training dataset. The learned \PDIshorts are effective at representing surfaces which are partially reflective or transparent. Figure~\ref{fig:discussion} (a) and (b) show two examples of such surfaces, rendered as anaglyphs with stereo-magnification.

Our method has certain limitations. When fine detail appears in front of a complex background, our model can struggle to place it at the correct depth. Figure~\ref{fig:discussion} (c) shows a case where overhead cables appear to jump between two different depths. This may suggest that depth decisions are being made too locally. Figure~\ref{fig:discussion} (d) shows the result of extrapolating beyond the limits of the \PDIshort representation. When the disparity between adjacent layers exceeds one pixel we may see duplicated edges, producing a ``stack of cards'' effect. 

In conclusion, we presented a new  representation, training setup, and approach to learning view extrapolation from video data. We believe this framework can also generalize to a variety of different tasks, including extrapolating from more than two input images or from only one, and generating lightfields allowing view movement in multiple dimensions.



\bibliographystyle{ACM-Reference-Format}
\bibliography{refs}


\begin{thebibliography}{55}


\ifx \showCODEN    \undefined \def \showCODEN     #1{\unskip}     \fi
\ifx \showDOI      \undefined \def \showDOI       #1{#1}\fi
\ifx \showISBNx    \undefined \def \showISBNx     #1{\unskip}     \fi
\ifx \showISBNxiii \undefined \def \showISBNxiii  #1{\unskip}     \fi
\ifx \showISSN     \undefined \def \showISSN      #1{\unskip}     \fi
\ifx \showLCCN     \undefined \def \showLCCN      #1{\unskip}     \fi
\ifx \shownote     \undefined \def \shownote      #1{#1}          \fi
\ifx \showarticletitle \undefined \def \showarticletitle #1{#1}   \fi
\ifx \showURL      \undefined \def \showURL       {\relax}        \fi
\providecommand\bibfield[2]{#2}
\providecommand\bibinfo[2]{#2}
\providecommand\natexlab[1]{#1}
\providecommand\showeprint[2][]{arXiv:#2}

\bibitem[\protect\citeauthoryear{Abadi, Barham, Chen, Chen, Davis, Dean, Devin,
  Ghemawat, Irving, Isard, et~al\mbox{.}}{Abadi et~al\mbox{.}}{2016}]%
        {abadi2016tensorflow}
\bibfield{author}{\bibinfo{person}{Mart{\'\i}n Abadi}, \bibinfo{person}{Paul
  Barham}, \bibinfo{person}{Jianmin Chen}, \bibinfo{person}{Zhifeng Chen},
  \bibinfo{person}{Andy Davis}, \bibinfo{person}{Jeffrey Dean},
  \bibinfo{person}{Matthieu Devin}, \bibinfo{person}{Sanjay Ghemawat},
  \bibinfo{person}{Geoffrey Irving}, \bibinfo{person}{Michael Isard},
  {et~al\mbox{.}}} \bibinfo{year}{2016}\natexlab{}.
\newblock \showarticletitle{TensorFlow: A System for Large-Scale Machine
  Learning}. In \bibinfo{booktitle}{\emph{OSDI}}.
\newblock


\bibitem[\protect\citeauthoryear{Agarwal, Mierle, and Others}{Agarwal
  et~al\mbox{.}}{2016}]%
        {ceres-solver}
\bibfield{author}{\bibinfo{person}{Sameer Agarwal}, \bibinfo{person}{Keir
  Mierle}, {and} \bibinfo{person}{Others}.} \bibinfo{year}{2016}\natexlab{}.
\newblock \bibinfo{title}{Ceres Solver}.
\newblock \bibinfo{howpublished}{\url{http://ceres-solver.org}}.
  (\bibinfo{year}{2016}).
\newblock


\bibitem[\protect\citeauthoryear{Apple}{Apple}{2016}]%
        {appleportraitmode}
\bibfield{author}{\bibinfo{person}{Apple}.} \bibinfo{year}{2016}\natexlab{}.
\newblock \bibinfo{title}{Portrait mode now available on i{P}hone 7 {P}lus with
  i{OS} 10.1}.
\newblock
  \bibinfo{howpublished}{\url{https://www.apple.com/newsroom/2016/10/portrait-mode-now-available-on-iphone-7-plus-with-ios-101/}}.
    (\bibinfo{year}{2016}).
\newblock


\bibitem[\protect\citeauthoryear{Ba, Kiros, and Hinton}{Ba
  et~al\mbox{.}}{2016}]%
        {ba2016layer}
\bibfield{author}{\bibinfo{person}{Jimmy~Lei Ba}, \bibinfo{person}{Jamie~Ryan
  Kiros}, {and} \bibinfo{person}{Geoffrey~E Hinton}.}
  \bibinfo{year}{2016}\natexlab{}.
\newblock \showarticletitle{Layer normalization}.
\newblock \bibinfo{journal}{\emph{arXiv preprint arXiv:1607.06450}}
  (\bibinfo{year}{2016}).
\newblock


\bibitem[\protect\citeauthoryear{Chapiro, Heinzle, Ayd{\i}n, Poulakos, Zwicker,
  Smolic, and Gross}{Chapiro et~al\mbox{.}}{2014}]%
        {chapiro2014optimizing}
\bibfield{author}{\bibinfo{person}{Alexandre Chapiro}, \bibinfo{person}{Simon
  Heinzle}, \bibinfo{person}{Tun{\c{c}}~Ozan Ayd{\i}n}, \bibinfo{person}{Steven
  Poulakos}, \bibinfo{person}{Matthias Zwicker}, \bibinfo{person}{Aljosa
  Smolic}, {and} \bibinfo{person}{Markus Gross}.}
  \bibinfo{year}{2014}\natexlab{}.
\newblock \showarticletitle{Optimizing stereo-to-multiview conversion for
  autostereoscopic displays}. In \bibinfo{booktitle}{\emph{Computer graphics
  forum}}.
\newblock


\bibitem[\protect\citeauthoryear{Chen, Papandreou, Kokkinos, Murphy, and
  Yuille}{Chen et~al\mbox{.}}{2018}]%
        {chen2018deeplab}
\bibfield{author}{\bibinfo{person}{Liang-Chieh Chen}, \bibinfo{person}{George
  Papandreou}, \bibinfo{person}{Iasonas Kokkinos}, \bibinfo{person}{Kevin
  Murphy}, {and} \bibinfo{person}{Alan~L Yuille}.}
  \bibinfo{year}{2018}\natexlab{}.
\newblock \showarticletitle{Deeplab: Semantic image segmentation with deep
  convolutional nets, atrous convolution, and fully connected crfs}.
\newblock \bibinfo{journal}{\emph{IEEE Trans. on Pattern Analysis and Machine
  Intelligence}} \bibinfo{volume}{40}, \bibinfo{number}{4}
  (\bibinfo{year}{2018}).
\newblock


\bibitem[\protect\citeauthoryear{Chen and Koltun}{Chen and Koltun}{2017}]%
        {chen2017photographic}
\bibfield{author}{\bibinfo{person}{Qifeng Chen} {and} \bibinfo{person}{Vladlen
  Koltun}.} \bibinfo{year}{2017}\natexlab{}.
\newblock \showarticletitle{Photographic image synthesis with cascaded
  refinement networks}. In \bibinfo{booktitle}{\emph{ICCV}}.
\newblock


\bibitem[\protect\citeauthoryear{Chen and Williams}{Chen and Williams}{1993}]%
        {chen1992view}
\bibfield{author}{\bibinfo{person}{Shenchang~Eric Chen} {and}
  \bibinfo{person}{Lance Williams}.} \bibinfo{year}{1993}\natexlab{}.
\newblock \showarticletitle{View Interpolation for Image Synthesis}. In
  \bibinfo{booktitle}{\emph{Proc. SIGGRAPH}}.
\newblock


\bibitem[\protect\citeauthoryear{Debevec, Taylor, and Malik}{Debevec
  et~al\mbox{.}}{1996}]%
        {debevec1996modeling}
\bibfield{author}{\bibinfo{person}{Paul~E Debevec}, \bibinfo{person}{Camillo~J
  Taylor}, {and} \bibinfo{person}{Jitendra Malik}.}
  \bibinfo{year}{1996}\natexlab{}.
\newblock \showarticletitle{Modeling and rendering architecture from
  photographs: A hybrid geometry-and image-based approach}. In
  \bibinfo{booktitle}{\emph{Proc. SIGGRAPH}}.
\newblock


\bibitem[\protect\citeauthoryear{Didyk, Sitthi-Amorn, Freeman, Durand, and
  Matusik}{Didyk et~al\mbox{.}}{2013}]%
        {didyk2013joint}
\bibfield{author}{\bibinfo{person}{Piotr Didyk}, \bibinfo{person}{Pitchaya
  Sitthi-Amorn}, \bibinfo{person}{William Freeman}, \bibinfo{person}{Fr{\'e}do
  Durand}, {and} \bibinfo{person}{Wojciech Matusik}.}
  \bibinfo{year}{2013}\natexlab{}.
\newblock \showarticletitle{Joint view expansion and filtering for
  automultiscopic 3D displays}. In \bibinfo{booktitle}{\emph{Proc. SIGGRAPH}}.
\newblock


\bibitem[\protect\citeauthoryear{Dosovitskiy and Brox}{Dosovitskiy and
  Brox}{2016}]%
        {dosovitskiy2016generating}
\bibfield{author}{\bibinfo{person}{Alexey Dosovitskiy} {and}
  \bibinfo{person}{Thomas Brox}.} \bibinfo{year}{2016}\natexlab{}.
\newblock \showarticletitle{Generating images with perceptual similarity
  metrics based on deep networks}. In \bibinfo{booktitle}{\emph{NIPS}}.
\newblock


\bibitem[\protect\citeauthoryear{Engel, Koltun, and Cremers}{Engel
  et~al\mbox{.}}{2018}]%
        {engel2018direct}
\bibfield{author}{\bibinfo{person}{Jakob Engel}, \bibinfo{person}{Vladlen
  Koltun}, {and} \bibinfo{person}{Daniel Cremers}.}
  \bibinfo{year}{2018}\natexlab{}.
\newblock \showarticletitle{Direct sparse odometry}.
\newblock \bibinfo{journal}{\emph{IEEE Trans. on Pattern Analysis and Machine
  Intelligence}} \bibinfo{volume}{40}, \bibinfo{number}{3}
  (\bibinfo{year}{2018}).
\newblock


\bibitem[\protect\citeauthoryear{Flynn, Neulander, Philbin, and Snavely}{Flynn
  et~al\mbox{.}}{2016}]%
        {flynn2016deepstereo}
\bibfield{author}{\bibinfo{person}{John Flynn}, \bibinfo{person}{Ivan
  Neulander}, \bibinfo{person}{James Philbin}, {and} \bibinfo{person}{Noah
  Snavely}.} \bibinfo{year}{2016}\natexlab{}.
\newblock \showarticletitle{Deep{S}tereo: {L}earning to Predict New Views From
  the World's Imagery}. In \bibinfo{booktitle}{\emph{CVPR}}.
\newblock


\bibitem[\protect\citeauthoryear{Forster, Pizzoli, and Scaramuzza}{Forster
  et~al\mbox{.}}{2014}]%
        {forster2014svo}
\bibfield{author}{\bibinfo{person}{Christian Forster}, \bibinfo{person}{Matia
  Pizzoli}, {and} \bibinfo{person}{Davide Scaramuzza}.}
  \bibinfo{year}{2014}\natexlab{}.
\newblock \showarticletitle{{SVO}: {F}ast Semi-Direct Monocular Visual
  Odometry}. In \bibinfo{booktitle}{\emph{ICRA}}.
\newblock


\bibitem[\protect\citeauthoryear{Garg and Reid}{Garg and Reid}{2016}]%
        {garg2016unsupervised}
\bibfield{author}{\bibinfo{person}{Ravi Garg} {and} \bibinfo{person}{Ian
  Reid}.} \bibinfo{year}{2016}\natexlab{}.
\newblock \showarticletitle{Unsupervised CNN for Single View Depth Estimation:
  Geometry to the Rescue}. In \bibinfo{booktitle}{\emph{ECCV}}.
\newblock


\bibitem[\protect\citeauthoryear{Godard, {Mac Aodha}, and Brostow}{Godard
  et~al\mbox{.}}{2017}]%
        {godard2016unsupervised}
\bibfield{author}{\bibinfo{person}{Cl{\'{e}}ment Godard},
  \bibinfo{person}{Oisin {Mac Aodha}}, {and} \bibinfo{person}{Gabriel~J.
  Brostow}.} \bibinfo{year}{2017}\natexlab{}.
\newblock \showarticletitle{Unsupervised Monocular Depth Estimation with
  Left-Right Consistency}. In \bibinfo{booktitle}{\emph{CVPR}}.
\newblock


\bibitem[\protect\citeauthoryear{Google}{Google}{2017a}]%
        {vr180}
\bibfield{author}{\bibinfo{person}{Google}.} \bibinfo{year}{2017}\natexlab{a}.
\newblock \bibinfo{title}{Introducing {VR}180 cameras}.
\newblock \bibinfo{howpublished}{\url{https://vr.google.com/vr180/}}.
  (\bibinfo{year}{2017}).
\newblock


\bibitem[\protect\citeauthoryear{Google}{Google}{2017b}]%
        {googleportrait}
\bibfield{author}{\bibinfo{person}{Google}.} \bibinfo{year}{2017}\natexlab{b}.
\newblock \bibinfo{title}{Portrait mode on the {P}ixel 2 and {P}ixel 2 {XL}
  smartphones}.
\newblock
  \bibinfo{howpublished}{\url{https://research.googleblog.com/2017/10/portrait-mode-on-pixel-2-and-pixel-2-xl.html}}.
    (\bibinfo{year}{2017}).
\newblock


\bibitem[\protect\citeauthoryear{Gortler, Grzeszczuk, Szeliski, and
  Cohen}{Gortler et~al\mbox{.}}{1996}]%
        {gortler1996lumigraph}
\bibfield{author}{\bibinfo{person}{Steven~J. Gortler}, \bibinfo{person}{Radek
  Grzeszczuk}, \bibinfo{person}{Richard Szeliski}, {and}
  \bibinfo{person}{Michael~F. Cohen}.} \bibinfo{year}{1996}\natexlab{}.
\newblock \showarticletitle{The Lumigraph}. In \bibinfo{booktitle}{\emph{Proc.
  SIGGRAPH}}.
\newblock


\bibitem[\protect\citeauthoryear{Ha, Im, Park, Jeon, and Kweon}{Ha
  et~al\mbox{.}}{2016}]%
        {ha2016high}
\bibfield{author}{\bibinfo{person}{Hyowon Ha}, \bibinfo{person}{Sunghoon Im},
  \bibinfo{person}{Jaesik Park}, \bibinfo{person}{Hae-Gon Jeon}, {and}
  \bibinfo{person}{In~So Kweon}.} \bibinfo{year}{2016}\natexlab{}.
\newblock \showarticletitle{High-quality Depth from Uncalibrated Small Motion
  Clip}. In \bibinfo{booktitle}{\emph{CVPR}}.
\newblock


\bibitem[\protect\citeauthoryear{Hartley and Zisserman}{Hartley and
  Zisserman}{2003}]%
        {hartley2003multiple}
\bibfield{author}{\bibinfo{person}{Richard Hartley} {and}
  \bibinfo{person}{Andrew Zisserman}.} \bibinfo{year}{2003}\natexlab{}.
\newblock \bibinfo{booktitle}{\emph{Multiple View Geometry in Computer
  Vision}}.
\newblock \bibinfo{publisher}{Cambridge University Press}.
\newblock


\bibitem[\protect\citeauthoryear{Hasinoff, Sharlet, Geiss, Adams, Barron,
  Kainz, Chen, and Levoy}{Hasinoff et~al\mbox{.}}{2016}]%
        {hasinoff2016burst}
\bibfield{author}{\bibinfo{person}{Samuel~W. Hasinoff}, \bibinfo{person}{Dillon
  Sharlet}, \bibinfo{person}{Ryan Geiss}, \bibinfo{person}{Andrew Adams},
  \bibinfo{person}{Jonathan~T. Barron}, \bibinfo{person}{Florian Kainz},
  \bibinfo{person}{Jiawen Chen}, {and} \bibinfo{person}{Marc Levoy}.}
  \bibinfo{year}{2016}\natexlab{}.
\newblock \showarticletitle{Burst photography for high dynamic range and
  low-light imaging on mobile cameras}. In \bibinfo{booktitle}{\emph{Proc.
  SIGGRAPH Asia}}.
\newblock


\bibitem[\protect\citeauthoryear{Hedman, Alsisan, Szeliski, and Kopf}{Hedman
  et~al\mbox{.}}{2017}]%
        {hedman2017casual}
\bibfield{author}{\bibinfo{person}{Peter Hedman}, \bibinfo{person}{Suhib
  Alsisan}, \bibinfo{person}{Richard Szeliski}, {and} \bibinfo{person}{Johannes
  Kopf}.} \bibinfo{year}{2017}\natexlab{}.
\newblock \showarticletitle{{Casual 3D Photography}}. In
  \bibinfo{booktitle}{\emph{Proc. SIGGRAPH Asia}}.
\newblock


\bibitem[\protect\citeauthoryear{Holroyd, Baran, Lawrence, and Matusik}{Holroyd
  et~al\mbox{.}}{2011}]%
        {holroyd2011multilayer}
\bibfield{author}{\bibinfo{person}{Michael Holroyd}, \bibinfo{person}{Ilya
  Baran}, \bibinfo{person}{Jason Lawrence}, {and} \bibinfo{person}{Wojciech
  Matusik}.} \bibinfo{year}{2011}\natexlab{}.
\newblock \showarticletitle{Computing and fabricating multilayer models}. In
  \bibinfo{booktitle}{\emph{Proc. SIGGRAPH Asia}}.
\newblock


\bibitem[\protect\citeauthoryear{Jaderberg, Simonyan, Zisserman,
  et~al\mbox{.}}{Jaderberg et~al\mbox{.}}{2015}]%
        {jaderberg2015spatial}
\bibfield{author}{\bibinfo{person}{Max Jaderberg}, \bibinfo{person}{Karen
  Simonyan}, \bibinfo{person}{Andrew Zisserman}, {et~al\mbox{.}}}
  \bibinfo{year}{2015}\natexlab{}.
\newblock \showarticletitle{Spatial transformer networks}. In
  \bibinfo{booktitle}{\emph{NIPS}}.
\newblock


\bibitem[\protect\citeauthoryear{Johnson, Alahi, and Fei-Fei}{Johnson
  et~al\mbox{.}}{2016}]%
        {johnson2016perceptual}
\bibfield{author}{\bibinfo{person}{Justin Johnson}, \bibinfo{person}{Alexandre
  Alahi}, {and} \bibinfo{person}{Li Fei-Fei}.} \bibinfo{year}{2016}\natexlab{}.
\newblock \showarticletitle{Perceptual losses for real-time style transfer and
  super-resolution}. In \bibinfo{booktitle}{\emph{ECCV}}.
\newblock


\bibitem[\protect\citeauthoryear{Kalantari, Wang, and Ramamoorthi}{Kalantari
  et~al\mbox{.}}{2016}]%
        {kalantari2016learning}
\bibfield{author}{\bibinfo{person}{Nima~Khademi Kalantari},
  \bibinfo{person}{Ting-Chun Wang}, {and} \bibinfo{person}{Ravi Ramamoorthi}.}
  \bibinfo{year}{2016}\natexlab{}.
\newblock \showarticletitle{Learning-Based View Synthesis for Light Field
  Cameras}. In \bibinfo{booktitle}{\emph{Proc. SIGGRAPH Asia}}.
\newblock


\bibitem[\protect\citeauthoryear{Kellnhofer, Didyk, Wang, Sitthi-Amorn,
  Freeman, Durand, and Matusik}{Kellnhofer et~al\mbox{.}}{2017}]%
        {kellnhofer20173dtv}
\bibfield{author}{\bibinfo{person}{Petr Kellnhofer}, \bibinfo{person}{Piotr
  Didyk}, \bibinfo{person}{Szu-Po Wang}, \bibinfo{person}{Pitchaya
  Sitthi-Amorn}, \bibinfo{person}{William Freeman}, \bibinfo{person}{Fredo
  Durand}, {and} \bibinfo{person}{Wojciech Matusik}.}
  \bibinfo{year}{2017}\natexlab{}.
\newblock \showarticletitle{{3DTV} at Home: {E}ulerian-{L}agrangian
  Stereo-to-Multiview Conversion}. In \bibinfo{booktitle}{\emph{Proc.
  SIGGRAPH}}.
\newblock


\bibitem[\protect\citeauthoryear{Kingma and Ba}{Kingma and Ba}{2014}]%
        {kingma2014adam}
\bibfield{author}{\bibinfo{person}{Diederik Kingma} {and}
  \bibinfo{person}{Jimmy Ba}.} \bibinfo{year}{2014}\natexlab{}.
\newblock \showarticletitle{Adam: {A} method for stochastic optimization}.
\newblock \bibinfo{journal}{\emph{arXiv preprint arXiv:1412.6980}}
  (\bibinfo{year}{2014}).
\newblock


\bibitem[\protect\citeauthoryear{Levoy and Hanrahan}{Levoy and
  Hanrahan}{1996}]%
        {levoy1996lightfield}
\bibfield{author}{\bibinfo{person}{Marc Levoy} {and} \bibinfo{person}{Pat
  Hanrahan}.} \bibinfo{year}{1996}\natexlab{}.
\newblock \showarticletitle{Light Field Rendering}. In
  \bibinfo{booktitle}{\emph{Proc. SIGGRAPH}}.
\newblock


\bibitem[\protect\citeauthoryear{Liu, Yeh, Tang, Liu, and Agarwala}{Liu
  et~al\mbox{.}}{2017}]%
        {liu2017video}
\bibfield{author}{\bibinfo{person}{Ziwei Liu}, \bibinfo{person}{Raymond Yeh},
  \bibinfo{person}{Xiaoou Tang}, \bibinfo{person}{Yiming Liu}, {and}
  \bibinfo{person}{Aseem Agarwala}.} \bibinfo{year}{2017}\natexlab{}.
\newblock \showarticletitle{Video Frame Synthesis Using Deep Voxel Flow}. In
  \bibinfo{booktitle}{\emph{ICCV}}.
\newblock


\bibitem[\protect\citeauthoryear{Lytro}{Lytro}{2018}]%
        {lytro}
\bibfield{author}{\bibinfo{person}{Lytro}.} \bibinfo{year}{2018}\natexlab{}.
\newblock \bibinfo{title}{Lytro}.
\newblock \bibinfo{howpublished}{\url{https://www.lytro.com/}}.
  (\bibinfo{year}{2018}).
\newblock


\bibitem[\protect\citeauthoryear{Mur-Artal and Tard\'os}{Mur-Artal and
  Tard\'os}{2015}]%
        {murartal2015orbslam}
\bibfield{author}{\bibinfo{person}{Montiel J. M.~M. Mur-Artal, Ra\'ul} {and}
  \bibinfo{person}{Juan~D. Tard\'os}.} \bibinfo{year}{2015}\natexlab{}.
\newblock \showarticletitle{{ORB-SLAM}: a Versatile and Accurate Monocular
  {SLAM} System}.
\newblock \bibinfo{journal}{\emph{IEEE Trans. on Robotics}}
  \bibinfo{volume}{31}, \bibinfo{number}{5} (\bibinfo{year}{2015}).
\newblock


\bibitem[\protect\citeauthoryear{Penner and Zhang}{Penner and Zhang}{2017}]%
        {penner2017soft3d}
\bibfield{author}{\bibinfo{person}{Eric Penner} {and} \bibinfo{person}{Li
  Zhang}.} \bibinfo{year}{2017}\natexlab{}.
\newblock \showarticletitle{Soft 3D Reconstruction for View Synthesis}. In
  \bibinfo{booktitle}{\emph{Proc. SIGGRAPH Asia}}.
\newblock


\bibitem[\protect\citeauthoryear{Porter and Duff}{Porter and Duff}{1984}]%
        {porter1984compositing}
\bibfield{author}{\bibinfo{person}{Thomas Porter} {and} \bibinfo{person}{Tom
  Duff}.} \bibinfo{year}{1984}\natexlab{}.
\newblock \showarticletitle{Compositing Digital Images}. In
  \bibinfo{booktitle}{\emph{Proc. SIGGRAPH}}.
\newblock


\bibitem[\protect\citeauthoryear{Riechert, Zilly, Kauff, G\"{u}ther, and
  Sch\"{a}fer}{Riechert et~al\mbox{.}}{2012}]%
        {riechert2012stereo}
\bibfield{author}{\bibinfo{person}{Christian Riechert},
  \bibinfo{person}{Frederik Zilly}, \bibinfo{person}{Peter Kauff},
  \bibinfo{person}{Jens G\"{u}ther}, {and} \bibinfo{person}{Ralf Sch\"{a}fer}.}
  \bibinfo{year}{2012}\natexlab{}.
\newblock \showarticletitle{Fully automatic stereo-to-multiview conversion in
  autostereoscopic displays}.
\newblock \bibinfo{journal}{\emph{The Best of IET and IBC}}
  \bibinfo{volume}{4} (\bibinfo{date}{09} \bibinfo{year}{2012}).
\newblock


\bibitem[\protect\citeauthoryear{Sch\"{o}nberger and Frahm}{Sch\"{o}nberger and
  Frahm}{2016}]%
        {schoenberger2016sfm}
\bibfield{author}{\bibinfo{person}{Johannes~Lutz Sch\"{o}nberger} {and}
  \bibinfo{person}{Jan-Michael Frahm}.} \bibinfo{year}{2016}\natexlab{}.
\newblock \showarticletitle{Structure-from-Motion Revisited}. In
  \bibinfo{booktitle}{\emph{CVPR}}.
\newblock


\bibitem[\protect\citeauthoryear{Shade, Gortler, He, and Szeliski}{Shade
  et~al\mbox{.}}{1998}]%
        {shade1998layered}
\bibfield{author}{\bibinfo{person}{Jonathan Shade}, \bibinfo{person}{Steven
  Gortler}, \bibinfo{person}{Li-wei He}, {and} \bibinfo{person}{Richard
  Szeliski}.} \bibinfo{year}{1998}\natexlab{}.
\newblock \showarticletitle{Layered depth images}. In
  \bibinfo{booktitle}{\emph{Proc. SIGGRAPH}}.
\newblock


\bibitem[\protect\citeauthoryear{Simonyan and Zisserman}{Simonyan and
  Zisserman}{2014}]%
        {simonyan2014very}
\bibfield{author}{\bibinfo{person}{Karen Simonyan} {and}
  \bibinfo{person}{Andrew Zisserman}.} \bibinfo{year}{2014}\natexlab{}.
\newblock \showarticletitle{Very deep convolutional networks for large-scale
  image recognition}.
\newblock \bibinfo{journal}{\emph{arXiv preprint arXiv:1409.1556}}
  (\bibinfo{year}{2014}).
\newblock


\bibitem[\protect\citeauthoryear{Srinivasan, Wang, Sreelal, Ramamoorthi, and
  Ng}{Srinivasan et~al\mbox{.}}{2017}]%
        {pratul2017lightfield}
\bibfield{author}{\bibinfo{person}{Pratul~P. Srinivasan},
  \bibinfo{person}{Tongzhou Wang}, \bibinfo{person}{Ashwin Sreelal},
  \bibinfo{person}{Ravi Ramamoorthi}, {and} \bibinfo{person}{Ren Ng}.}
  \bibinfo{year}{2017}\natexlab{}.
\newblock \showarticletitle{Learning to Synthesize a 4D RGBD Light Field from a
  Single Image}. In \bibinfo{booktitle}{\emph{ICCV}}.
\newblock


\bibitem[\protect\citeauthoryear{Tulsiani, Zhou, Efros, and Malik}{Tulsiani
  et~al\mbox{.}}{2017}]%
        {tulsiani2017multiview}
\bibfield{author}{\bibinfo{person}{Shubham Tulsiani}, \bibinfo{person}{Tinghui
  Zhou}, \bibinfo{person}{Alexei~A. Efros}, {and} \bibinfo{person}{Jitendra
  Malik}.} \bibinfo{year}{2017}\natexlab{}.
\newblock \showarticletitle{Multi-view Supervision for Single-view
  Reconstruction via Differentiable Ray Consistency}. In
  \bibinfo{booktitle}{\emph{CVPR}}.
\newblock


\bibitem[\protect\citeauthoryear{Vijayanarasimhan, Ricco, Schmid, Sukthankar,
  and Fragkiadaki}{Vijayanarasimhan et~al\mbox{.}}{2017}]%
        {vijayanarasimhan2017sfmnet}
\bibfield{author}{\bibinfo{person}{Sudheendra Vijayanarasimhan},
  \bibinfo{person}{Susanna Ricco}, \bibinfo{person}{Cordelia Schmid},
  \bibinfo{person}{Rahul Sukthankar}, {and} \bibinfo{person}{Katerina
  Fragkiadaki}.} \bibinfo{year}{2017}\natexlab{}.
\newblock \showarticletitle{Sfm-net: Learning of structure and motion from
  video}.
\newblock \bibinfo{journal}{\emph{arXiv preprint arXiv:1704.07804}}
  (\bibinfo{year}{2017}).
\newblock


\bibitem[\protect\citeauthoryear{Wang and Adelson}{Wang and Adelson}{1994}]%
        {wang1994representing}
\bibfield{author}{\bibinfo{person}{John~YA Wang} {and}
  \bibinfo{person}{Edward~H Adelson}.} \bibinfo{year}{1994}\natexlab{}.
\newblock \showarticletitle{Representing moving images with layers}.
\newblock \bibinfo{journal}{\emph{IEEE Trans. on Image Processing}}
  \bibinfo{volume}{3}, \bibinfo{number}{5} (\bibinfo{year}{1994}).
\newblock


\bibitem[\protect\citeauthoryear{Wang, Bovik, Sheikh, and Simoncelli}{Wang
  et~al\mbox{.}}{2004}]%
        {wang2004image}
\bibfield{author}{\bibinfo{person}{Zhou Wang}, \bibinfo{person}{Alan Bovik},
  \bibinfo{person}{Hamid Sheikh}, {and} \bibinfo{person}{Eero Simoncelli}.}
  \bibinfo{year}{2004}\natexlab{}.
\newblock \showarticletitle{Image quality assessment: from error visibility to
  structural similarity}.
\newblock \bibinfo{journal}{\emph{IEEE Trans. on Image Processing}}
  \bibinfo{volume}{13}, \bibinfo{number}{4} (\bibinfo{year}{2004}).
\newblock


\bibitem[\protect\citeauthoryear{Wanner, Meister, and Goldluecke}{Wanner
  et~al\mbox{.}}{2013}]%
        {wanner2013datasets}
\bibfield{author}{\bibinfo{person}{Sven Wanner}, \bibinfo{person}{Stephan
  Meister}, {and} \bibinfo{person}{Bastian Goldluecke}.}
  \bibinfo{year}{2013}\natexlab{}.
\newblock \showarticletitle{Datasets and benchmarks for densely sampled 4d
  light fields}. In \bibinfo{booktitle}{\emph{VMV}}.
\newblock


\bibitem[\protect\citeauthoryear{Wetzstein, Lanman, Heidrich, and
  Raskar}{Wetzstein et~al\mbox{.}}{2011}]%
        {wetzstein2011layered}
\bibfield{author}{\bibinfo{person}{G. Wetzstein}, \bibinfo{person}{D. Lanman},
  \bibinfo{person}{W. Heidrich}, {and} \bibinfo{person}{R. Raskar}.}
  \bibinfo{year}{2011}\natexlab{}.
\newblock \showarticletitle{{Layered 3D: Tomographic Image Synthesis for
  Attenuation-based Light Field and High Dynamic Range Displays}}. In
  \bibinfo{booktitle}{\emph{Proc. SIGGRAPH}}.
\newblock


\bibitem[\protect\citeauthoryear{Wikipedia}{Wikipedia}{2017}]%
        {wikimultiplanecamera}
\bibfield{author}{\bibinfo{person}{Wikipedia}.}
  \bibinfo{year}{2017}\natexlab{}.
\newblock \bibinfo{title}{Multiplane camera}.
\newblock
  \bibinfo{howpublished}{\url{https://en.wikipedia.org/wiki/Multiplane_camera}}.
    (\bibinfo{year}{2017}).
\newblock


\bibitem[\protect\citeauthoryear{Xie, Girshick, and Farhadi}{Xie
  et~al\mbox{.}}{2016}]%
        {xie2016deep3d}
\bibfield{author}{\bibinfo{person}{Junyuan Xie}, \bibinfo{person}{Ross~B.
  Girshick}, {and} \bibinfo{person}{Ali Farhadi}.}
  \bibinfo{year}{2016}\natexlab{}.
\newblock \showarticletitle{Deep3{D}: {F}ully Automatic 2{D}-to-3{D} Video
  Conversion with Deep Convolutional Neural Networks}. In
  \bibinfo{booktitle}{\emph{ECCV}}.
\newblock


\bibitem[\protect\citeauthoryear{Yu and Gallup}{Yu and Gallup}{2014}]%
        {yu2014accidental}
\bibfield{author}{\bibinfo{person}{Fisher Yu} {and} \bibinfo{person}{David
  Gallup}.} \bibinfo{year}{2014}\natexlab{}.
\newblock \showarticletitle{3D Reconstruction from Accidental Motion}. In
  \bibinfo{booktitle}{\emph{CVPR}}.
\newblock


\bibitem[\protect\citeauthoryear{Yu and Koltun}{Yu and Koltun}{2016}]%
        {yu2015multi}
\bibfield{author}{\bibinfo{person}{Fisher Yu} {and} \bibinfo{person}{Vladlen
  Koltun}.} \bibinfo{year}{2016}\natexlab{}.
\newblock \showarticletitle{Multi-Scale Context Aggregation by Dilated
  Convolutions}. In \bibinfo{booktitle}{\emph{ICLR}}.
\newblock


\bibitem[\protect\citeauthoryear{Zhang, Isola, Efros, Shechtman, and
  Wang}{Zhang et~al\mbox{.}}{2018}]%
        {zhang2018perceptual}
\bibfield{author}{\bibinfo{person}{Richard Zhang}, \bibinfo{person}{Phillip
  Isola}, \bibinfo{person}{Alexei~A Efros}, \bibinfo{person}{Eli Shechtman},
  {and} \bibinfo{person}{Oliver Wang}.} \bibinfo{year}{2018}\natexlab{}.
\newblock \showarticletitle{The Unreasonable Effectiveness of Deep Networks as
  a Perceptual Metric}. In \bibinfo{booktitle}{\emph{CVPR}}.
\newblock


\bibitem[\protect\citeauthoryear{Zhang, Liu, and Dai}{Zhang
  et~al\mbox{.}}{2015}]%
        {zhang2015light}
\bibfield{author}{\bibinfo{person}{Zhoutong Zhang}, \bibinfo{person}{Yebin
  Liu}, {and} \bibinfo{person}{Qionghai Dai}.} \bibinfo{year}{2015}\natexlab{}.
\newblock \showarticletitle{Light field from micro-baseline image pair}. In
  \bibinfo{booktitle}{\emph{CVPR}}.
\newblock


\bibitem[\protect\citeauthoryear{Zhou, Brown, Snavely, and Lowe}{Zhou
  et~al\mbox{.}}{2017}]%
        {zhou2017unsupervised}
\bibfield{author}{\bibinfo{person}{Tinghui Zhou}, \bibinfo{person}{Matthew
  Brown}, \bibinfo{person}{Noah Snavely}, {and} \bibinfo{person}{David Lowe}.}
  \bibinfo{year}{2017}\natexlab{}.
\newblock \showarticletitle{Unsupervised learning of depth and ego-motion from
  video}. In \bibinfo{booktitle}{\emph{CVPR}}.
\newblock


\bibitem[\protect\citeauthoryear{Zhou, Tulsiani, Sun, Malik, and Efros}{Zhou
  et~al\mbox{.}}{2016}]%
        {zhou2016view}
\bibfield{author}{\bibinfo{person}{Tinghui Zhou}, \bibinfo{person}{Shubham
  Tulsiani}, \bibinfo{person}{Weilun Sun}, \bibinfo{person}{Jitendra Malik},
  {and} \bibinfo{person}{Alexei~A Efros}.} \bibinfo{year}{2016}\natexlab{}.
\newblock \showarticletitle{View synthesis by appearance flow}. In
  \bibinfo{booktitle}{\emph{ECCV}}.
\newblock


\bibitem[\protect\citeauthoryear{Zitnick, Kang, Uyttendaele, Winder, and
  Szeliski}{Zitnick et~al\mbox{.}}{2004}]%
        {zitnick2004interpolation}
\bibfield{author}{\bibinfo{person}{C.~Lawrence Zitnick},
  \bibinfo{person}{Sing~Bing Kang}, \bibinfo{person}{Matthew Uyttendaele},
  \bibinfo{person}{Simon Winder}, {and} \bibinfo{person}{Richard Szeliski}.}
  \bibinfo{year}{2004}\natexlab{}.
\newblock \showarticletitle{High-quality Video View Interpolation Using a
  Layered Representation}. In \bibinfo{booktitle}{\emph{Proc. SIGGRAPH}}.
\newblock


\end{thebibliography}

\end{document}